\def\year{2021}\relax
\documentclass[letterpaper]{article} 
\usepackage{aaai21}  
\usepackage{times}  
\usepackage{helvet} 
\usepackage{courier}  
\usepackage[hyphens]{url}  
\usepackage{graphicx} 
\usepackage{natbib}  
\usepackage{caption} 
\usepackage{enumitem}   
\usepackage{amssymb,amsmath,amsthm, mathtools}
\usepackage{algorithm}
\usepackage{algorithmic}

\def\st{\medspace|\medspace}
\def\woe{\text{woe}}
\def\cB{\mathcal{B}}
\def\R{\mathbb{R}}

\DeclareMathOperator*{\argmax}{argmax}
\newcommand\tightdots{\hbox to 1em{.\hss.\hss.}}

\usepackage[dvipsnames,usenames]{xcolor}

\newcommand{\edit}[1]{\textcolor{black}{#1}}

\title{From Human Explanation to Model Interpretability:\\ A Framework Based on Weight of Evidence}
\author{
    David Alvarez-Melis,\textsuperscript{\rm 1}
    Harmanpreet Kaur,\textsuperscript{\rm 2} \\
    Hal Daum\'{e} III,\textsuperscript{\rm 1, \rm 3}
    Hanna Wallach,\textsuperscript{\rm 1}
    Jennifer Wortman Vaughan\textsuperscript{\rm 1} \\
}
\affiliations{
    \textsuperscript{\rm 1} Microsoft Research \\
    \textsuperscript{\rm 2} University of Michigan \\
    \textsuperscript{\rm 3} University of Maryland\\
    \{alvarez.melis,hal3,hanna,jenn\}@microsoft.com, harmank@umich.edu

}

\begin{document}

\maketitle

\begin{abstract}
  We take inspiration from the study of human explanation to inform the design and evaluation of interpretability methods \edit{in machine learning}. First, we survey the literature on human explanation in philosophy, cognitive science, and the social sciences, and propose a list of design principles for machine-generated explanations that are meaningful to humans. Using the concept of weight of evidence from information theory, we develop a method for \edit{generating} explanations that adhere to these principles.  We show that this method can be adapted to handle high-dimensional, multi-class settings, yielding a flexible framework for generating explanations. We demonstrate that these explanations can be estimated accurately from finite samples and are robust to small perturbations of the inputs. We also evaluate our method through a qualitative user study with machine learning practitioners, where we observe that the resulting explanations are usable despite some participants struggling with background concepts like prior class probabilities. Finally, we conclude by surfacing~design~implications for interpretability tools in general.\looseness=-1
\end{abstract}

\section{Introduction}
\label{sec:intro}

Interpretability has long been a desirable property of machine learning (ML) models. With the success of complex models like neural networks, and their expanding reach into high-stakes and decision-critical applications, explaining ML models' predictions has become even more important. Interpretability can enable model debugging and lead to more robust ML systems, support knowledge discovery, and boost trust~\cite{HHB20}. It can also help to mitigate unfairness by surfacing undesirable model behavior~\cite{TCHL18, dodge19explaining}, lead to increased accountability by enabling auditing~\cite{SB18}, and enable ML practitioners to better communicate model behavior to stakeholders~\cite{BVV+18,HHB20}.\looseness=-1

There are two primary techniques for achieving interpretability of ML models. The first is to train transparent, or \emph{glass-box}, models that are intended to be inherently interpretable, such as decision trees \citep{quinlan1986induction} and sets \citep{lakkaraju2016interpretable}, simple point systems~\cite{zeng2017interpretable,jung2020simple}, and generalized additive models~\cite{hastie1990generalized,Caruana2015-qf}.  Although some researchers have argued that glass-box models should always be used in high-stakes scenarios~\cite{rudin2018please}, complex \emph{black-box} models, such as neural networks, random forests, and ensemble methods, are very widely used in practice. As a result, other ML researchers have gravitated towards interpretability methods that generate post-hoc \emph{local} explanations for individual predictions produced by~such models~\citep[e.g.,][]{simonyan2013deep, selvaraju2017grad, ribeiro2016why, lundberg2017unified, alvarez-melis2017causal}.\looseness=-1

Local explanations aim to answer the question of why a model $\mathcal{M}$ predicted a particular output $y$ for some input $x$. There are many ways of operationalizing this abstract question, but most methods \edit{do so by addressing} the proxy question of how much the value of each input feature $x_i$ contributed to the prediction $y$. Thus, in practice, the explanations generated by many such methods consist of importance scores indicating the positive or negative relevance of each input feature $x_i$. Although the way these scores are computed varies from method to method, most start from an axiomatic or algorithmic derivation of some notion of feature importance, and only later investigate whether the resulting explanations are useful to humans. Some methods forgo this last step altogether, relying exclusively on intrinsic evaluation of mathematical properties of explanations, such as robustness~or faithfulness to~the~underlying model~\citep{alvarez-melis2018robustness, alvarez-melis2018towards, jacovi2020towards}.\looseness=-1

Interpretability, however, is fundamentally a human-centered concept. In \edit{light} of this, we put human needs at the center of both the design and evaluation of interpretability methods.
Our work builds upon and weaves together two literatures that study the relationship between humans and explanations. First, researchers in philosophy, cognitive science, and the social sciences have long studied what it means to explain, and how humans do it \citep[e.g.,][and references therein]{pitt1988theories,miller2019explanation}. Second, a recent line of work within the human--computer interaction community has focused on how humans understand and utilize interpretability tools (i.e., software implementations of interpretability methods)~\cite{LD11,BLL12,BSO15, hohman2019gamut,LCH+19,AvdW+20,PG+21}, where a common finding is that practitioners misunderstand, over-trust,~and~misuse~these tools \citep[e.g.,][]{kaur2020interpreting}.\looseness=-1

Inspired by \citet{miller2019explanation}, we start by surveying the literature on the nature of explanation, revealing recurring characteristics of human explanation that are often missing from interpretability methods. We distill these characteristics into design principles that we argue human-centered machine-generated explanations should satisfy.
We then realize our design principles \edit{using} the concept of \textit{weight of evidence} (WoE) from information theory \citep{good1985weight}, which has recently been advocated for by \citet{spiegelhalter2018neurips}, but, to the best of our knowledge, has yet to be investigated in the context of interpretability.
We demonstrate that WoE can be adapted to handle high-dimensional, multi-class settings, yielding a suitable theoretical foundation for interpretability. We provide a general, customizable meta-algorithm to generate explanations for black-box models. We also show experimentally that WoE can be estimated from finite samples and is robust to small perturbations of the inputs.\looseness=-1

Evaluation of interpretability methods is notoriously difficult ~\citep{doshi-velez2017towards,kaur2020interpreting}.  Although recent work has focused on abstract, intrinsic metrics such as robustness or faithfulness to the underlying model~\citep{alvarez-melis2018robustness}, considerably less attention has been given to understanding how the resulting explanations are used in practice.  This discrepancy between the intended use of the explanations\,---\,by a human, for a specific goal such as auditing, debugging, or building trust in a model\,---\,and their experimental evaluation\,---\,\edit{typically performed using} abstract\edit{, intrinsic} metrics, in generic settings\,---\,hampers understanding of the benefits~and failure points of different \edit{interpretability} methods.\looseness=-1

We build on a recent thread of work~\cite[e.g.,][]{LCH+19,nourani2019effects,li2020understanding,kaur2020interpreting,VW21,PG+21}\,---\,including several recent papers from the human computation community\,---\,that argues that evaluations should be grounded in concrete use cases and should put humans at the center, taking into account not only how they use interpretability tools, but how well they understand the principles behind them. We carry out an artifact-based interview study with ten ML practitioners to investigate their use of a tool implementing our meta-algorithm in the context of a practical task. Qualitative themes from this study suggest that most participants successfully used the tool to answer questions, despite struggling with background concepts like prior class probabilities. Although the study was designed to identify preferences for different tool modalities, participants often used all of them and requested the option to switch between them interactively. Our results additionally highlight the importance of providing well-designed tutorials for interpretability tools\,---\,even for experienced ML practitioners\,---\,which are often overlooked in the literature \edit{on interpretability methods}, and which we argue should be an integral part of any interpretability tool.\looseness=-1

\section{Human-Centered Design Principles}\label{sec:design_principles}
What it means to explain and how humans do it have long been studied in philosophy, cognitive science, and the social sciences. We draw on this literature to propose human-centered design principles for interpretability methods.

\citet{hempel1948studies} and \citet{vanfraassen1988pragmatic} define an explanation as consisting of two main pieces: the \textit{explanandum}, a description of the phenomenon to be explained, and the \textit{explanans}, the facts or propositions that explain the phenomenon, which may rely on relevant aspects of context.
As is often done colloquially, we will refer to the explanans as the \emph{explanation}. Different ways of formalizing the explanation have given rise to various theories, ranging from logical deterministic propositions \citep{hempel1948studies} to probabilistic ones \citep{salmon1971statistical, vanfraassen1988pragmatic}. An excellent historical overview can be~found in the surveys by~\citet{pitt1988theories} and~\citet{miller2019explanation}.

In the context of local explanations for predictions made by ML models, the phenomenon to be explained is why a model $\mathcal{M}$ predicted output $y$ for input $x$. This why-question can be operationalized in different ways.  The facts used to explain this phenomenon may include information about the input features, the model parameters, the data used to train the model, or the manner in which the model was trained.

Although the nature of explanation is far from settled, recurring themes emerge across disciplines. At the core of the theories by \citet{vanfraassen1988pragmatic} and \citet{lipton1990contrastive} is the hypothesis that humans tend to explain in contrastive terms  (e.g., ``a fever is more consistent with the flu than with a cold''), with explanations that are both factual and counterfactual (e.g., ``had the patient had chest pressure too, the diagnosis would instead have been bronchitis''). Yet, the explanations produced by most current interpretability methods refer only to why the input $x$ points to a single hypothesis (i.e., the prediction $y$) rather than ruling out all alternatives.\footnote{Exceptions include recent work advocating for contrastive or counterfactual explanations~\citep{wachter2017counterfactual, miller2019contrastive, vanderwaa2018contrastive}, partly inspired by contrast sets \citep{azevedo2010rules, bay1999detecting, webb2003detecting, novak2009supervised}.} In light of this, we propose our first two design principles:
\begin{enumerate}[topsep=2pt, leftmargin=1.5em,noitemsep]
    \item \textbf{Explanations should be contrastive}, i.e., explicate why the model predicted $y$ instead of alternative $y'$.
    \item \textbf{Explanations should be exhaustive}, i.e., provide a justification for why every alternative $y'$ was not predicted.
\end{enumerate}

Another theme, featured prominently by \citet{hempel1962deductive}, is that human explanations decompose into simple components. In other words, humans usually explain using multiple simple accumulative statements, each addressing a few aspects of the evidence (e.g., ``a fever rules out a cold in favor of bronchitis or pneumonia; among these, chills suggest the latter''). Each component is intended to be understood without further decomposition. Again, this contrasts with current interpretability methods that explain in one shot, for example, by providing importance scores for all features simultaneously. Our next two design principles are therefore:\looseness=-1
\begin{enumerate}[topsep=2pt, leftmargin=1.5em,noitemsep]
    \setcounter{enumi}{2}
    \item \textbf{Explanations should be modular and compositional}, breaking up predictions into simple components.
    \item \textbf{Explanations should rely on easily-understandable quantities}, so that each component is understandable.
\end{enumerate}

Another recurring theme is minimality. In a survey of over 250 papers,
\citet{miller2019explanation} argued that it is important, but underappreciated in ML, that only the most relevant facts be included in explanations.
Thus, our final principle is:\looseness=-1
\begin{enumerate}[topsep=2pt, leftmargin=1.5em,noitemsep]
    \setcounter{enumi}{4}
    \item  \textbf{Explanations should be parsimonious}, i.e., only the most relevant facts should be provided as components.
\end{enumerate}

These design principles are not exhaustive; each could be refined or generalized, and other principles could be derived from the same literature. However, we posit that these principles provide a reasonable starting point because they capture some of the most apparent discrepancies between human and machine-generated explanations. More generally, these principles point to a broader theme of human explanations as a \textit{process} rather than (only) a \textit{product} \citep{miller2019explanation, lombrozo2012explanation}. Therefore, these principles work to shift interpretability methods from the latter towards the former.

\section{Explaining with the Weight of Evidence}\label{sec:woe}
The set of design principles proposed in the previous section outlines a framework for human-centered interpretability in ML. In this section, we show how this framework can be operationalized by means of the \emph{weight of evidence}, a simple but powerful concept from information theory. We operationalize the question of why model $\mathcal{M}$ predicted output $y$ for input $x$ in terms of how much \emph{evidence} each input feature $x_i$ (or feature group) provides in favor of $y$ relative to alternatives.  An explanation based on this question adheres to our design principles because it is based on a familiar concept (evidence) that is grounded in common language, it naturally evokes a contrastive statement (evidence \textit{for} or \textit{against} something), and, as we explain below, it can be formalized using simple quantities that admit modularity.\looseness=-1

\label{sec:background}
\subsection{Weight of Evidence: Foundations}\label{sec:woe_foundations}

The weight of evidence (WoE) is a well-studied probabilistic approach for analyzing variable importance that traces its origins back to \citet{pierce1878probability}, but was popularized by \citet{good1950probability, good1968corroboration, good1985weight}, whose definition and notation we follow here.
Given a hypothesis and some evidence, the WoE seeks to answer the following question:
\textit{"How much does the~evidence speak in favor of or against the hypothesis?"}\looseness=-1

The \edit{WoE} is usually \edit{defined} for some evidence $e$, a hypothesis $h$, and its logical complement $\overline{h}$. For example, in a simple binary classification setting\edit{,} $e=(X_1, \dots, X_n)$, $h: Y=1$, and $\overline{h}: Y=0$. The WoE of $e$ in favor of $h$ is the log-odds ratio between $h$ conditioned on $e$ and $h$ marginally:\looseness=-1
\begin{equation}\label{eq:woe_def}
    \text{woe}(h : e) \triangleq  \log \frac{O( h \mid e) }{O(h)},
\end{equation}
where $O(\cdot)$ denotes the odds of a hypothesis, i.e.,
\begin{equation}
    O(h) \triangleq \frac{P(h)}{P(\overline{h})}\qquad \text{and} \qquad O( h \mid e) \triangleq\frac{P(h \mid e)}{P(\overline{h} \mid e)}.
\end{equation}
Using Bayes' rule, $\woe(h : e)$ can also be defined as
\begin{equation}\label{eq:woe_alter_def}
    \woe(h : e) \triangleq \log \frac{P(e \mid h)}{P(e\mid \overline{h})}.
\end{equation}
These two equivalent definitions provide complementary views of the WoE: the \textit{hypothesis-odds} and \textit{evidence-likelihood} interpretations. Using Equation~\eqref{eq:woe_def}, $\woe(h:e) > 0$ indicates that the odds of $h$ are higher under $e$ than marginally. Equivalently, using Equation~\eqref{eq:woe_alter_def}, it indicates that the likelihood of $e$ is larger when conditioning on $h$ than on its complement. In other words, the evidence \textit{speaks in favor of} hypothesis $h$. Analogously, if $\text{woe}(h:e) < 0 $ we would say that the evidence \textit{speaks against} $h$. The quantities in Equations~\eqref{eq:woe_def} and \eqref{eq:woe_alter_def} are contrastive (cf.~Principle 1)\,---\,that is, defined in terms of ratios.\looseness=-1

As a concrete example, suppose that a doctor wants to know whether a patient's symptoms indicate the presence of a certain disease, say, the flu. Denote $e=$ ``\textit{the patient has a fever},'' $h=$ ``\textit{the patient has the flu},'' and $\bar{h}=$ ``\textit{the patient doesn't have the flu}.'' The doctor might know that for a patient, the odds of having the flu roughly double once the patient's fever is taken into account (i.e., the hypothesis-odds interpretation), which corresponds to $\text{woe}(h:e) \approx \log 2$. Alternatively, using the evidence-likelihood interpretation, the doctor could conclude that a patient is twice as likely to have a fever if they have the flu compared to when they do not. Note that neither interpretation tells us anything about the base rate of the flu.\looseness=-1

The WoE generalizes beyond these simple scenarios. For example, it can be conditioned on additional information $c$:
\[
\text{woe}(h : e \mid c) \triangleq  \log \frac{P( e \mid h, c) }{P(e \mid \overline{h}, c)} .
\]
It can also contrast $h$ to an arbitrary alternative hypothesis $h'$ instead of $\bar{h}$  (e.g., evidence in favor of \edit{the flu} and against \edit{a cold}): $\text{woe}(h/h': e) \triangleq  \text{woe}(h: e \mid h \lor h')$. Thus, we can, in general, talk about the strength of evidence in favor of $h$ and against $h'$ provided by $e$ (perhaps conditioned on $c$).\looseness=-1

When the evidence is decomposable into multiple parts\,---\,that is, when $e= \bigcup_{i=1}^n e_i$\,---\,the WoE is also decomposable:
\begin{equation}\label{eq:woe_multievidence}
    \woe(h/h': e) = \sum_{i=1}^n \log \frac{P(e_i \mid e_{i-1}, \dots, e_1, h)}{P(e_i \mid e_{i-1}, \dots, e_1, h')} .
\end{equation}
This is crucial to defining an extension of the WoE to high-dimensional inputs that adheres to Principle 3 (modularity).

A further appealing aspect of the WoE is its immediate connection to Bayes' rule through the following identity:\looseness=-1
\begin{equation}\label{eq:woe_prior_posterior}
   \underbrace{\log \frac{P(h \mid e)}{P(h' \mid  e)}}_{\text{Posterior log odds}}  =\underbrace{\log\frac{P(h)}{P(h')}}_{\text{Prior log odds}} + \underbrace{\log \frac{P(e \mid h)}{P(e \mid h')}}_{\text{Weight of evidence}}.
\end{equation}
In other words, the WoE can be understood as an adjustment \edit{to} the prior log odds caused by observing the evidence. \edit{In~a~simple binary classification setting, this amounts to}
\begin{equation*}
    \log \frac{P(Y\!=\!1 \mid\! X)}{P(Y\!=\!0 \mid \! X)}  =  \log\frac{P(Y\!=\!1)}{P(Y\!=\!0)} + \text{woe}(Y=1 : X), 
\end{equation*}
which shows that a positive (respectively, negative) WoE implies that the posterior log odds of $Y=1$ versus~$Y=0$ are higher (lower) than the prior log odds, indicating that the evidence~makes $Y=1$ more (less) likely than it was \textit{a priori}.

Equation~\eqref{eq:woe_prior_posterior} shows that the WoE is modular (cf.~Principle 3) in another important way: it disentangles prior class probabilities and input likelihoods. This is important because of the \textit{base rate fallacy} studied in the behavioral science literature \citep{tversky1974judgment, bar-hillel1980base, eddy1982probabilistic, koehler1996base}. This cognitive bias, prevalent even among domain experts, is characterized by a frequent misinterpretation of posterior probabilities, primarily caused by a neglect of base rates (i.e., prior probabilities). Despite this, many interpretability methods do not explicitly display prior probabilities, and even when they do, they focus on explaining posterior probabilities, which invariably entangle~information about priors and the input being explained.\looseness=-1

Additionally, the units in which the WoE is expressed (log-odds ratios) are arguably easily understandable (cf.~Principle 4). There is evidence from the cognitive-neuroscience literature that log odds are a natural unit in human cognition. For example\edit{,} degrees of confidence expressed by humans are proportional to log odds \citep{peirce1885small}\edit{,} people are less biased when responding in log odds that in linear scales \citep{phillips1966conservatism}, and there exist plausible neurological hypotheses for encoding of log odds in the human brain \citep{gold2001neural, gold2002banburisms}. We refer the reader to \citet{Zhang2012ubiquitous} for a meta-analysis of these various studies of log odds.\looseness=-1

We provide \edit{additional} properties of the WoE, along with an axiomatic derivation\edit{,} in the appendix, \edit{which can be found in the longer version of this paper, available online.}\footnote{\url{https://arxiv.org/abs/2104.13299}}\looseness=-1

\subsection{Composite Hypotheses and Evidence}\label{sec:woe_extension}
Traditionally, the WoE has been mostly used in simple settings, such as a single binary output and only a few input features. Its use in the more complex settings typically considered in modern ML therefore poses new challenges.\looseness=-1

The first such challenge is that in multi-class \edit{settings}, there is flexibility in choosing the hypotheses $h$ and $h'$ to contrast. The obvious choice of letting $h$ correspond to the predicted class $y^*$ and $h'$ its complement is unlikely to yield useful explanations when the number of classes is large (e.g., explaining the evidence in favor of one disease against one hundred thousand others). Following Principle 3 (modularity), and taking inspiration from Hempel's model \citeyearpar{hempel1962deductive} and the view of explanation as a process \citep{lombrozo2012explanation, miller2019explanation}, we address this by casting explanation as a sequential procedure, whereby a subset of \edit{the} possible classes is ruled out at each step. For example, in medical diagnosis, we might first explain why bacterial diseases were ruled out in favor of viral ones, and then explain why a specific viral disease was predicted instead of the others. In general, for a classification problem over labels $\mathsf{Y}=\{1,\dots,k\}$, we will consider a (given or constructed) nested partitioning of $\mathsf{Y}$ into a sequence of $T$-many subsets $\mathsf{U}_i$ of classes such that $\{y^*\} \triangleq \mathsf{U}_T \subset \mathsf{U}_{T-1} \subset \dots \subset \mathsf{U}_{0} \triangleq \mathsf{Y}$.  As \edit{we} show in \edit{Figure 4} in the appendix, this partition implies a sequence of pairs of hypotheses $(h_t,h_t')=(y\!\in\!\mathsf{U}_t,y\!\in\!\mathsf{U}_{t-1}\!\setminus\!\mathsf{U}_t)$.\looseness=-1

A second challenge arises when the the number of input features is large. For very high-dimensional inputs (such as images or detailed health records), providing a WoE value for each feature will rarely be informative. Again, imagine our hypothetical doctor having to simultaneously analyze the relevance of thousands of symptoms. For such cases, we propose aggregating the input features into feature groups (e.g., super-pixels for images or groups of related symptoms for medical diagnosis). Formally, for an input $X$ of dimension $n$, we partition the feature indices into $m$ disjoint subsets, with $\mathsf{S}_1 \cup \dots \cup \mathsf{S}_m = \{1,\dots,n\}$.  Equation~\eqref{eq:woe_multievidence} (or, equivalently, the chain rule of probability) allows for arbitrary groupings, so for any such partition we can compute
\vspace{-0.1cm}
\begin{equation}\label{eq:atom_woe}
 \!\text{woe}( h/h'\!:\!X) =\!\!\sum_{i=1}^m \underbrace{\log \frac{P(X_{\mathsf{S}_i}\!\mid\!X_{\mathsf{S}_{i-1}}, \tightdots, X_{\mathsf{S}_1}, h)}{P(X_{\mathsf{S}_i}\!\mid\!X_{\mathsf{S}_{i-1}}, \tightdots, X_{\mathsf{S}_1}, h')}}_{= \woe(h/h': X_{\mathsf{S}_i}\mid X_{\mathsf{S}_{i-1}}, \dots, X_{\mathsf{S}_1})}
\end{equation}
\vspace{-0.1cm}
where $X_{\mathsf{S}_i} = \{X_j\}_{j \in \mathsf{S}_i}$ is the $i$th feature group, or ``atom.''

\begin{algorithm}[t!]
  \caption{WoE meta-algorithm for complex models}\label{algo:greedy_woe}
\begin{algorithmic}[1]
   \STATE {\bfseries Input:} Instance $X \in \R^n$, prediction $y^* \in \{1,\dots, k\}$
   \STATE {\bfseries Parameters:} Features $\mathcal{A} = \{\{1\}, \dots,\{n\}\} $ or feature groups $\mathcal{A} = \{S_1, \dots, S_m\}$
   \STATE Initialize $\mathsf{U}_0 \gets \{1,\dots, k\}$ 
   \STATE $t \gets 0$\;
   \WHILE{$|\mathsf{U}_t| > 1$}
   \STATE $t \gets t+1$\;
   \STATE $\mathsf{U}_{t} \gets \textsc{SelectHypothesis}(\mathsf{U}_{t-1}, y^*)$\;
    \STATE $\overline{\mathsf{U}}_{t} \gets \mathsf{U}_{t-1} \setminus \mathsf{U}_{t}$ \COMMENT{relative complement}
    \STATE $\pi(\mathsf{U}_{t}) \gets \log \tfrac{P( y \in \mathsf{U}_{t})}{P(y \in \overline{\mathsf{U}}_{t})}$ \COMMENT{prior log odds}
   \FOR{$i=1,\dots, |\mathcal{A}|$}
    \STATE $\omega_i^t \!\gets\! \woe(y\!\in\!\mathsf{U}_{t}/y\!\in\!\overline{\mathsf{U}}_{t} \!:\! X_{\mathcal{A}_i} \!\mid \! X_{\mathcal{A}_{i-1}}, \!\dots,\! X_{\mathcal{A}_{1}})  $\;
   \ENDFOR
    \STATE $\Omega_t \gets \sum_{i=1}^{|\mathcal{A}|}\omega_i^t$
    \STATE $\textsc{DisplayExplanation}(\mathsf{U}_{t}, \overline{\mathsf{U}}_{t}, \mathcal{A}, \pi(\mathsf{U}_t), \{\omega_i^t\}_i, \Omega_t)$\;
   \ENDWHILE
\end{algorithmic}
\end{algorithm}

\begin{figure*}[htp!]
    \centering
    \includegraphics[width=0.5\linewidth]{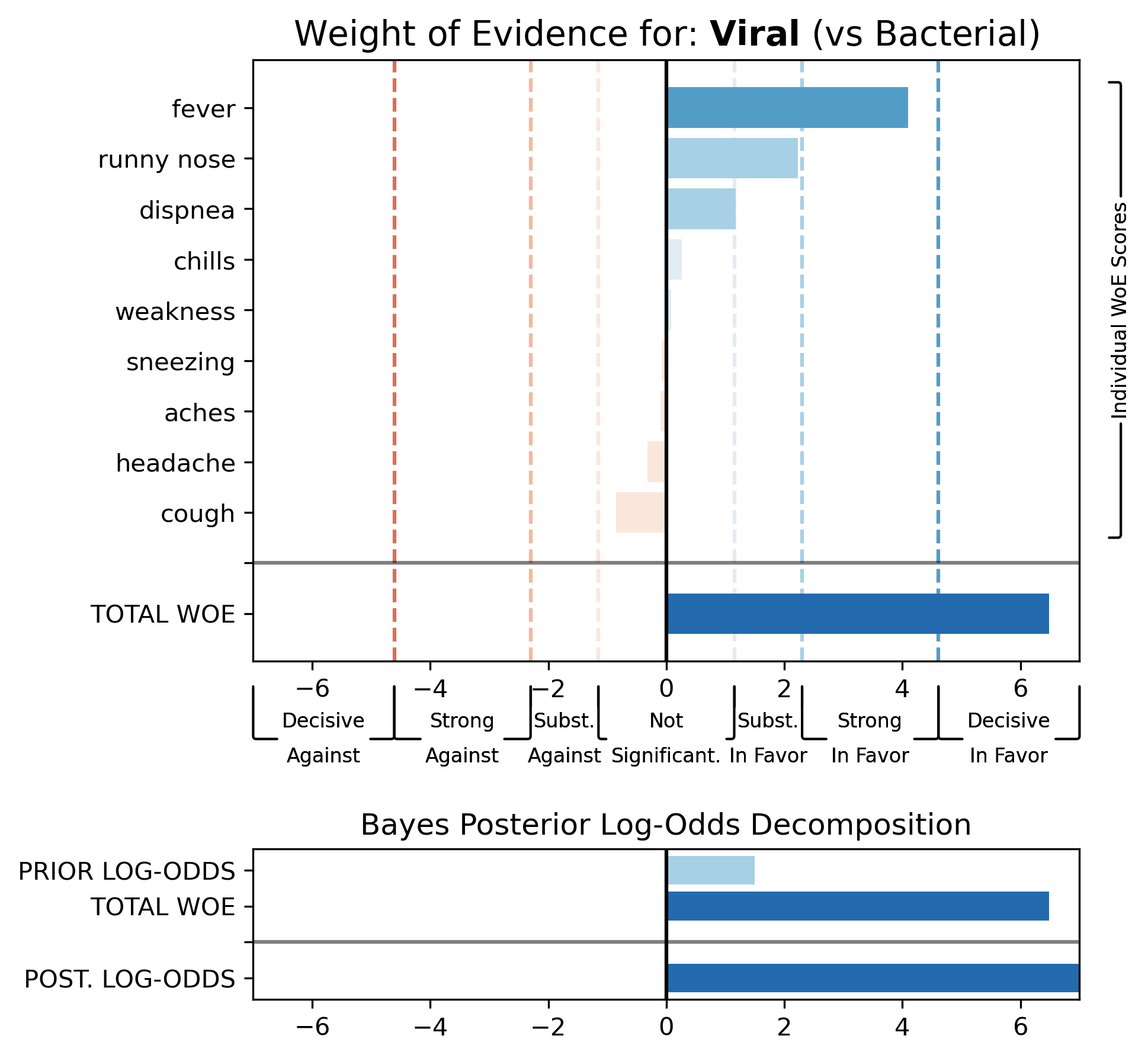}%
    \includegraphics[width=0.5\linewidth]{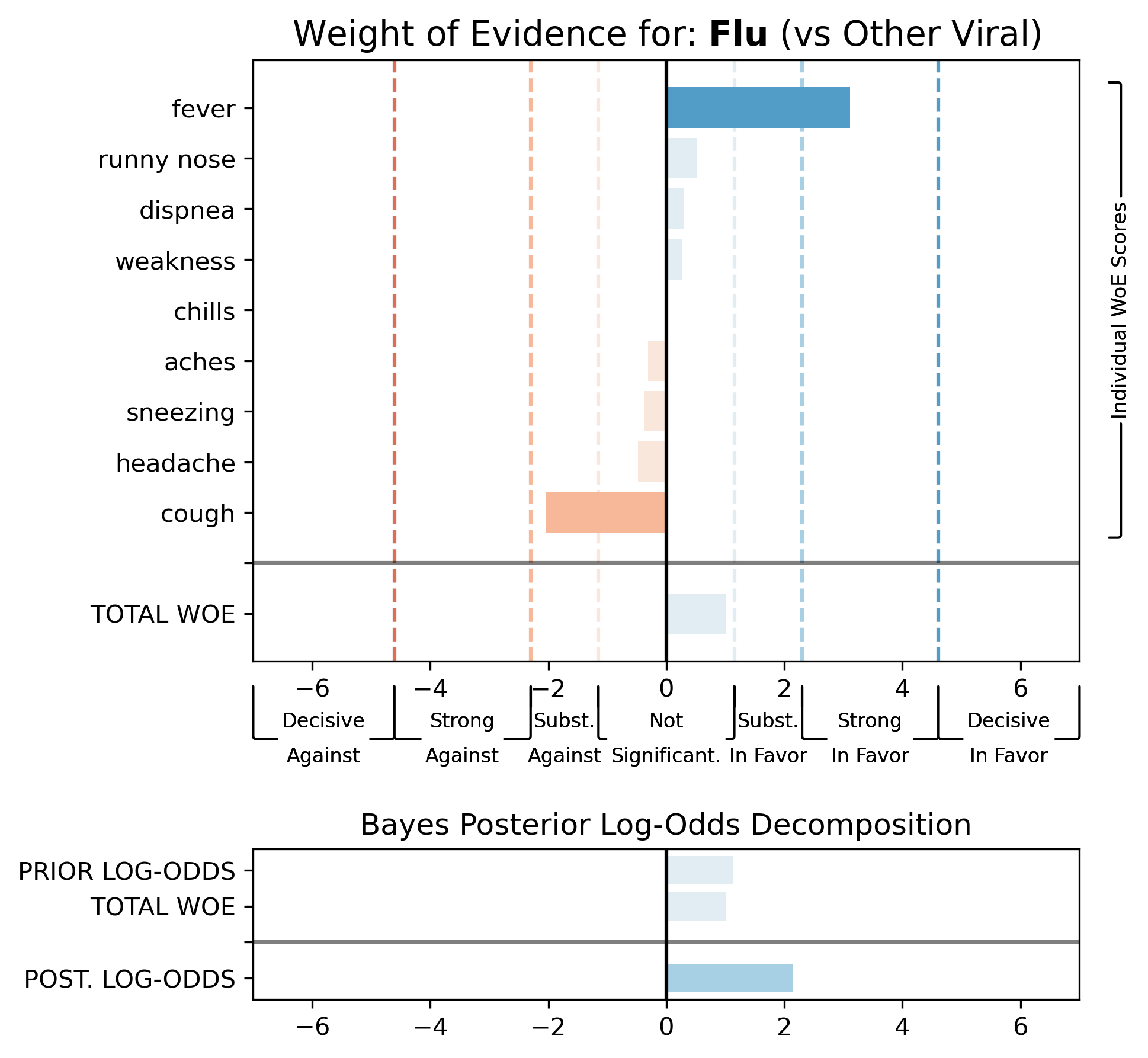}\\
    \caption{Example two-step explanation produced by our \edit{method} for a \edit{model that} predict\edit{ed} the \edit{flu for some input}. The first step \edit{explains}, using \edit{the} weight of evidence, why the model favors \edit{viral diseases} (instead of \edit{bacterial ones}) for this \edit{input}. Then, the second step explains why the model predict\edit{ed the flu instead of} the remaining possible classes (\edit{i.e.,} other \edit{viral} diseases).}\label{fig:example_expl}
\end{figure*}

\subsection{A Meta-Algorithm for WoE Explanations}\label{sec:woe_algo}

Using these extensions of the WoE, we propose a meta-algorithm for generating explanations for complex classifiers (Algorithm~\ref{algo:greedy_woe}). Given a model, an input, and a prediction, the algorithm generates an explanation for the prediction sequentially by producing WoE values for progressively smaller nested hypotheses. Specifically, at every step $t$, a subset of classes $\mathsf{U}_t \subset \mathsf{U}_{t-1}$ is selected and the remaining classes $\overline{\mathsf{U}}_{t} = \mathsf{U}_{t-1} \setminus \mathsf{U}_t$ \edit{are} ruled out.  The user is shown a comparison of hypotheses $h_t: y \in \mathsf{U}_t$ and $h_t': y \in \overline{\mathsf{U}}_{t}$ consisting of both their prior log odds $\pi(\mathsf{U}_{t})$ \edit{(line 9)} and the WoE in favor of $h_t$ and against $h_t'$.  WoE values are computed sequentially with each atom $\mathcal{A}_i$ (either an individual input feature or a group of features) as the evidence (line 11) and these values are summed to obtain the total WoE using the additive property (line 13). These values are presented to the user, and the process continues until all classes except the prediction $y^*$ have been ruled out (cf.~Principles 2--3).\looseness=-1

Left unspecified in this meta-algorithm are four key choices that are application-dependent and require further discussion.
First is the question of how to define the $\textsc{SelectHypothesis}$ method to progressively partition the classes (line 7). If there is an inherent natural partitioning (e.g., ``viral'' versus ``bacterial,'' as in the example discussed previously), then $\textsc{SelectHypothesis}$ simply amounts to retrieving the largest subset in the partition containing the prediction $y^*$. For the general case, we propose selecting~the~hypothesis that maximizes a WoE-based objective:
\[ \mathsf{U}_{t}\! \gets \!{\displaystyle \argmax_{\mathsf{U} \subset \mathsf{U}_{t-1}; y^*\in \mathsf{U}}} \woe(y \in \mathsf{U}\medspace/\medspace(y \in \mathsf{U}_{t-1}\!\setminus\!\mathsf{U}) : X) \!-\! R(\mathsf{U}), \]
where $R$ is a cardinality-based regularizer. $R$ should be chosen to penalize sets that are too small (which would yield granular explanations with many steps, in opposition to Principle 5) or too large (which would yield coarse explanations, to the detriment of Principle 3). Although the choice of $R$ should ideally be informed by the application and the user, a sensible generic choice is $R(\mathsf{U}) \propto \left||\mathsf{U}|-\frac{1}{2}|\mathsf{U}_{t-1}| \right|^p$, normalized so that $R(\mathsf{U})\in [0,1]$. Using this regularizer, Algorithm 1 approximately splits the remaining classes in half at every step, yielding roughly $O(\log k)$ steps in total.\looseness=-1

Second, it should be noted that lines 10--12 in Algorithm~\ref{algo:greedy_woe} implicitly assume an ordering of the atoms, and that this ordering might affect the WoE values. In some applications, there might be a conditional independence structure known a priori that could inform the choice of atoms and their ordering (e.g., those simplifying the conditioning in line 11 the most). If not, the ordering can again be chosen randomly or based on the sorted per-atom conditional WoE values.\looseness=-1

Third, computing the per-atom WoE (line 11) requires the conditional likelihoods  $P(X_{\mathcal{A}_i} | X_{\mathcal{A}_{i-1}}, \dots, X_{\mathcal{A}_1}, Y)$. Ideally, the \edit{model} would compute these likelihoods internally. If, instead, it computes only marginal feature likelihoods  $P(X_{\mathcal{A}_i} | Y)$, we can use a na\"ive Bayes (NB) approximation\,---\,that is, use these in place of the conditional likelihoods in Equation~\eqref{eq:atom_woe}. If the model is a black box or does not compute \edit{conditional} likelihoods internally,~then~these must be estimated \edit{as we explain below}. \looseness=-1

Finally, there is the question of how to implement $\textsc{DisplayExplanation}$. When the number of atoms is large, the WoE values for only the most salient atoms can be displayed (cf.~Principle 5)\,---\,e.g., those with absolute WoE larger than a given threshold $\tau$; \citet{good1985weight} suggests $\tau= 2$ as a rule of thumb. Otherwise, all per-atom WoE values can be displayed along with the total WoE and prior log odds.

An example two-step explanation produced by our method (on fabricated data from our user study tutorial) is shown in Figure~\ref{fig:example_expl}. The sorting and color coding of the features by their WoE value\edit{s} makes it apparent which of these contribute the most evidence in favor or against the selected class (or set of classes), and the labeling along the x-axis provides guidelines for context. The visualization suggests the additive nature of these values (i.e., that stacking blue bars and subtracting red ones yields the total WoE). The bottom panel, a graphical representation of Equation~\eqref{eq:woe_prior_posterior}, disentangles the model's estimated prior class odds (which \textit{a priori} weakly favor \edit{the} flu and other viral diseases), from its total WoE (very strong when contrasting \edit{viral and bacterial~diseases}, less so for the \edit{flu versus other viral diseases}).\looseness=-1

\subsection{WoE Estimation for Black-box Models}

As \edit{we} noted \edit{above}, computing \edit{the per-atom} WoE \edit{(line 11 in Algorithm 1)} requires evaluating the conditional likelihoods $P(X_{\mathcal{A}_i} | X_{\mathcal{A}_{i-1}}, \dots, X_{\mathcal{A}_1}, Y)$. In many practical settings, \edit{including those in which the model is a black box}, these \edit{conditional likelihoods are not computed internally}, so they \edit{must} be estimated. In such \edit{settings}, we propose \edit{fitting} a conditional likelihood \edit{estimation} model \edit{using the model's predictions $\hat{y}=\mathcal{M}(x)$ (not the true labels $y$) as a preliminary step. This conditional likelihood estimation model can then be called on demand when computing the per-atom WoE}.\looseness=-1

In some settings\edit{,} it \edit{may} be \edit{possible} to fit a full (i.e., conditioned
on \textit{both} the class $Y$ and all previous atoms \edit{$A_i-1, \ldots,
A_1$) conditional likelihood estimation model}, for example, via kernel
or spectral density estimation methods when \edit{working with}
low-dimensional data, or \edit{via} autoregressive or recurrent neural
networks \edit{when working with} text or time-series data. For more complex
\edit{types of data}, such as images, methods based on normalizing flows and
neural autoregressive models \citep[e.g.,][]{rezende2015variational,
papamakarios2017masked} are likely \edit{to be} more appropriate. \edit{In settings
where fitting a full conditional likelihood estimation model is
infeasible}, an NB approximation can be used \edit{to estimate} class- (but
not atom-) conditional likelihoods\edit{, for example,} via a Gaussian NB
classifier.\looseness=-1

We emphasize that fitting \edit{a conditional} likelihood estimation
model\,---\,the main computational bottleneck \edit{of} our method\,---\,must be done
only once, potentially offline. This is in contrast \edit{to}
perturbation-based interpretability methods~like LIME that fit \edit{a
new} model for every prediction.\looseness=-1

We \edit{assess} the quality of finite-sample WoE estimation experimentally
in the \edit{``}Quantitative Experiments\edit{''} section. \looseness=-1

\subsection{Relation to LIME and SHAP}

When viewed from a probabilistic perspective, most post-hoc interpretability methods revolve around a model's predictive posterior\,---\,that is, they seek explanations that deconstruct $P(Y=y^*\mid X)$ in various ways. For example, LIME \citep{ribeiro2016why} seeks to approximate $f(x) = P(Y \mid X=x)$ in the vicinity of $x_0$ through a simpler, interpretable surrogate model $\tilde{f}(x)$. Similarly, SHAP \citep{lundberg2017unified} quantifies variable importance by analyzing the effect on the posterior of ``dropping'' variables $X_i$ from the input $X$. In contrast, the WoE focuses\,---\,directly, in the case of the evidence-likelihood interpretation and indirectly in the case of the hypothesis-odds interpretation\,---\,on the conditional likelihood $P(X = x \mid Y)$. In other words, for a given input $x$, a WoE explanation is based on the probability assigned by the model to $x$ (or a subset of its features) given, e.g.,
$Y=y^*$.\looseness=-1

When viewed in this way, the relationship between WoE explanations and
other post-hoc interpretability methods like LIME and SHAP is akin to
the relationship between generative and discriminative models. Indeed,
as is the case for some pairs of generative and discriminative models
(e.g., na\"{i}ve Bayes and logistic regression), these different
interpretability methods also turn out to be equivalent\,---\,two sides of
the same coin\,---\,for some simple classifiers, as \edit{we} show in
the appendix for logistic regression. However,
this is not generally the case. Moreover, even when the explanations
generated by different interpretability methods qualitatively agree (i.e., the same
features are highlighted as being important), the specific
interpretations of the explanations will differ. Indeed, the WoE uses
a different operationalization of the notion of feature importance, in turn entailing different
units of explanation: log likelihoods and log\edit{-}odds ratios in the case
of the WoE and linear attribution scores for~posterior probabilities in the case of LIME and SHAP.\looseness=-1

\section{Quantitative Experiments}
\label{sec:quant}

Here, we assess the quality of finite-sample WoE estimation and the robustness of the WoE to perturbations of the inputs.\looseness=-1

\begin{figure}
    \centering
    \includegraphics[width=\linewidth, trim={0 1cm 0 0}, clip]{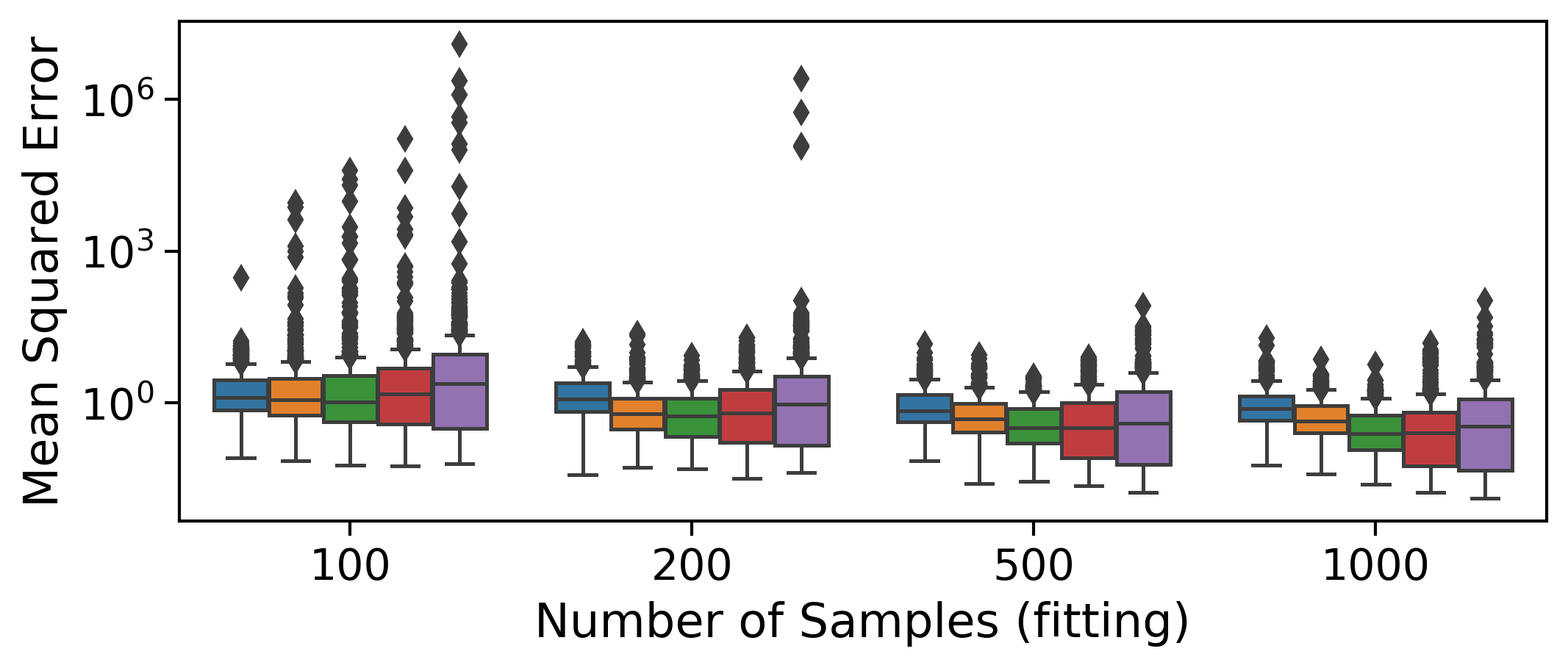}
    \includegraphics[width=\linewidth, trim={0 0.25cm 0 0}, clip]{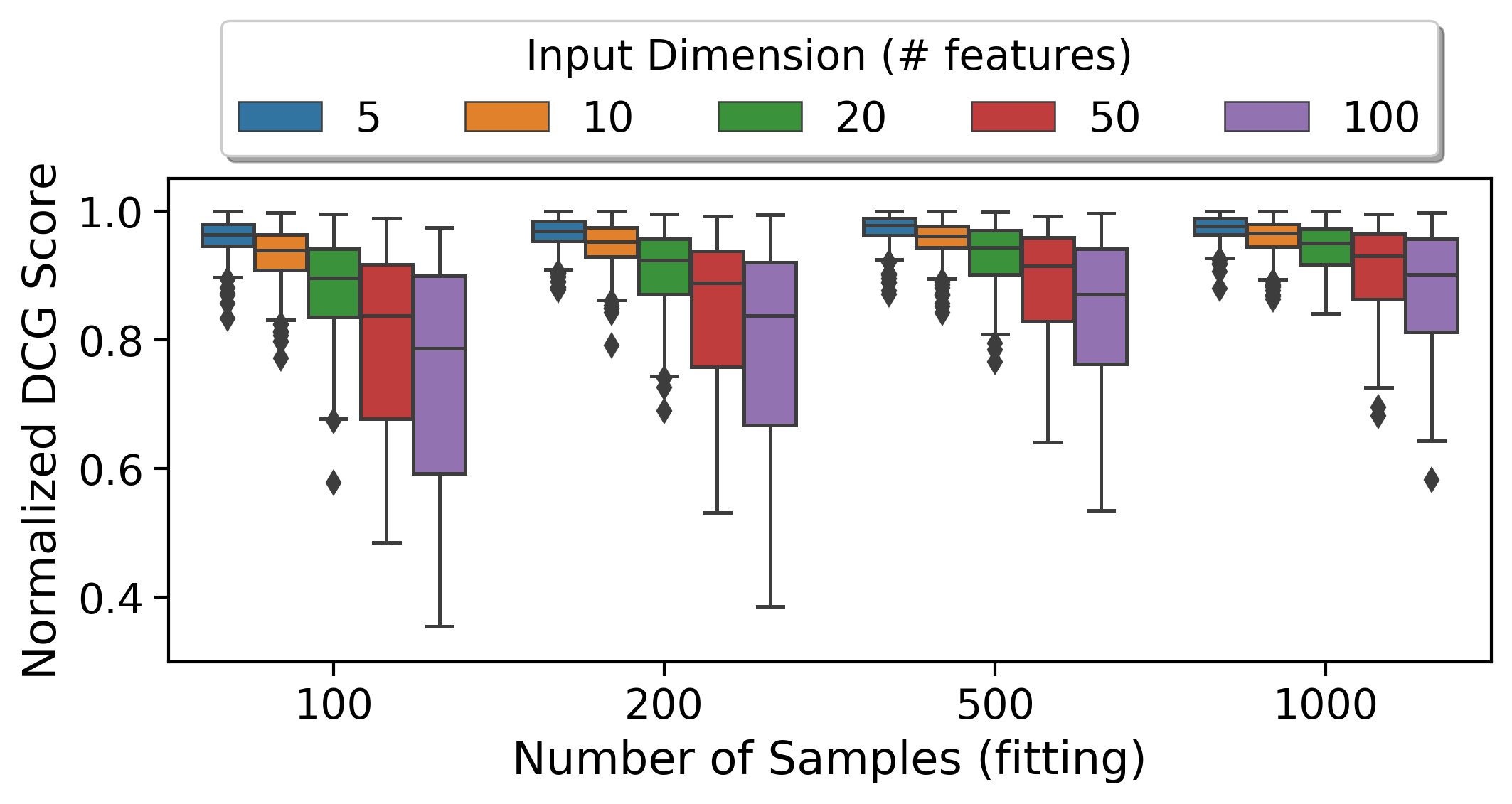}
    \caption{Quality of WoE estimation. Top: MSE. Bottom: NDCG, ranging from $0$ (worst) to $1$ (perfect) ranking quality.}
    \label{fig:recovery_plot}
\end{figure}

\begin{figure*}[ht!]
    \centering
    \includegraphics[width=0.9\linewidth, trim={0 1cm 0 0}]{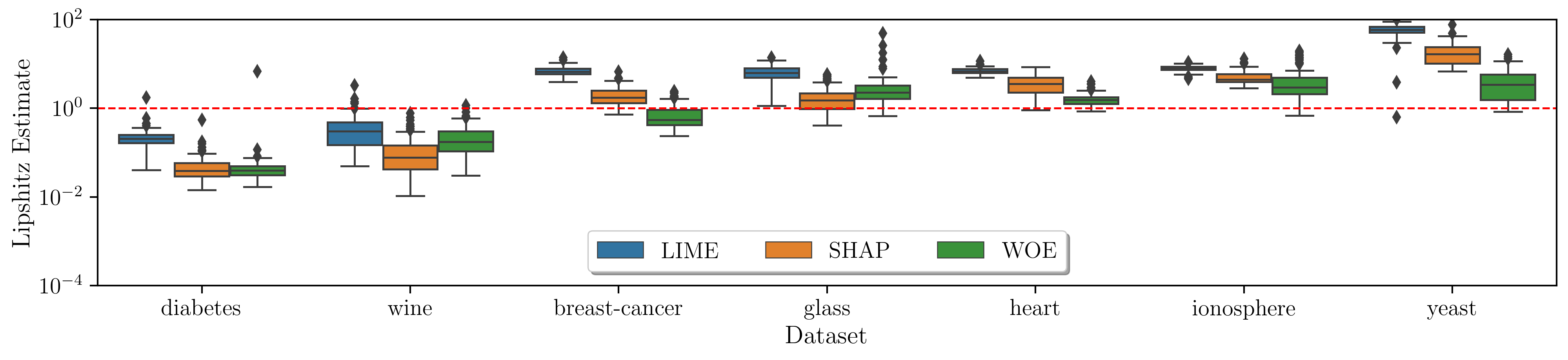}
    \caption{Explanation robustness across benchmark datasets. Extreme values away from $L=1$ (dashed line) are undesirable.}
    \label{fig:robustness_plot}
\end{figure*}

\subsection{Quality of Finite-Sample WoE Estimation}\label{sec:woe_approx_experiments}
Our first experiment evaluates the quality of WoE estimates from finite samples. As \edit{we} explained above, such estimates are needed if the model is a black box or does not compute likelihoods \edit{internally}. For evaluation purposes, we consider a model that, by construction, computes all the quantities required for exact WoE computation, but treat it as black box\,---\,that is, its internal WoE computation will be used only for evaluation, and is not available to \edit{our method}. Instead, we separately fit a \edit{conditional} likelihood estimation model by querying the model for a small number of inputs, and use this estimation model to compute WoE values at explanation time. We control for model \edit{mis}specification by having both the model and our \edit{conditional} likelihood estimation model use the \edit{NB} assumption. Specifically, we use a smoothed Gaussian \edit{NB} (GNB) classifier.
This allows us to focus on the quality of WoE estimation from finite~samples, but does not address model misspecification.\looseness=-1

First, we generate a dataset of a given dimension. We train a \edit{model} on a subset of this dataset of size $N_{\text{train}}=1000$ and fit the \edit{conditional} likelihood estimation model on a separate subset of size $N_{\text{fit}}$, which we vary. For every test \edit{input} $x_i$ ($N_{\text{test}}=10$), we compute true WoE values for each input feature using the model's prior and posterior probabilities, and then compute estimated WoE values according to the estimation model in the appendix.
We compare these using two metrics: mean squared error and normalized discounted cumulative gain (NDCG), a measure of ranking quality that might be relevant to practitioners, applied to the relative ranking of input features by their (true or estimated) WoE.\footnote{The NDCG is only defined for positive values, so we compute it separately for positive and negative values and average them.}\looseness=-1

Figure~\ref{fig:recovery_plot} shows these metrics as a function of
input dimension and sample size $N_{\text{fit}}$. As expected,
estimation quality improves with the number of samples used for
fitting, and degrades gracefully as the input dimension
increases. These results suggest that the WoE can be accurately
estimated\,---\,even in relatively high dimensions\,---\,from finite samples,
\edit{although} we caution that these results may look different for other
models, and we do not measure \edit{model misspecification} error due to the
NB approximation.\looseness=-1

\subsection{Robustness of WoE}\label{sec:woe_robustness_experiments}
Previous work has argued that interpretability methods should be robust in the sense that the explanations they provide should not vary dramatically when the input whose prediction is being explained changes by a small amount.  To investigate the robustness of our method, we follow the set-up of~\citet{alvarez-melis2018robustness}.  Letting $\mathcal{E}(\cdot)$ be a function that maps feature vectors $x\in\R^n$ to explanation vectors (e.g., importance scores) $e \in \R^n$, we quantify its robustness around $x_0$ through its local Lipschitz constant:
\begin{equation}\label{eq:lipschitz}
    L(x_0) = \max_{x_j \in \cB_{\varepsilon}(x_0)} \frac{\| \mathcal{E}(x_j) - \mathcal{E}(x_0)\|}{\|x_j - x_0\|} ,
\end{equation}
where $ \cB_{\varepsilon}(x_0) = \{ x \st \|x - x_0\| \leq \varepsilon\}$.  Intuitively, $L(x_0)$ quantifies the largest relative change in importance scores in a small neighborhood around $x_0$. Extreme values are usually undesirable, as they indicate explanations that are either too sensitive (large $L$) or not responsive enough (very small $L$) to changes in the input features. In most settings, values below $1$ but bounded away from $0$ are preferable.\looseness=-1

Concretely, we first train a GNB classifier\edit{.} Then, for any input $x$ we use the classifier to generate a prediction $y$, and input both of these to our method to generate $\mathcal{E}(x)$, a vector of WoE values for each feature $x_i$. Since computing the robustness metric \eqref{eq:lipschitz} involves maximization, we estimate this quantity from finite samples using Bayesian optimization, making repeated calls to $\mathcal{E}(\cdot)$. We focus on standard benchmark classification datasets from the UCI repository \citep{dua2017uci}, and compare \edit{our method to} LIME \citep{ribeiro2016why} and SHAP \citep{lundberg2017unified}. Figure~\ref{fig:robustness_plot} shows \edit{the} results \edit{for} ten repetitions with different random seeds for each \edit{dataset and} interpretability \edit{method} pair. The red dashed line indicates the bound $L(x)=1$. \edit{For} all but one dataset, WoE explanations are\edit{,} on average\edit{,} as close or closer to the ideal Lipschitz robustness as \edit{explanations} generated by LIME and SHAP.\looseness=-1

\section{User Study}\label{sec:user_study}
Throughout this paper, we have argued that interpretability is fundamentally a human-centered concept and that the evaluation of interpretability methods should therefore focus on the needs of humans, exploring how they use interpretability tools, as well as their understanding of the concepts that underlie them. In this section, we present a user study that we carried out to assess the usefulness of our method in the types of scenarios in which it would be used in practice. Such studies are commonly used in the HCI community to distinguish between designers' intended use of a tool and users' mental models~\cite{gibson1977theory,norman2013design}.\looseness=-1

\subsection{Study Design}

We conducted an artifact-based interview study with 10 ML practitioners to evaluate the use of our interpretability \edit{method, implemented in a simple} tool (the artifact)\edit{,} in a controlled setting. In such qualitative studies, the goal is to \edit{ensure sufficient interaction} time and nuanced data collection \edit{for each} participant, which is typically only feasible for relatively small sample sizes~\cite{hudson2014concepts,olsen2007evaluating,turner2006determining}.
\edit{Our} study followed a think-aloud protocol\edit{, in which participants were asked to verbalize their thought processes as they used the tool to perform specific tasks, in order to help us identify specific} concepts and \edit{functionalities} that \edit{might} be confusing. We place\edit{d} participants in a controlled setting \edit{(}rather than observe them using the tool on their own models\edit{)} because of challenges including data access,
inconsistencies in the types of data analyzed by the \edit{participants}, and \edit{the potential} difficulty \edit{of} establishing patterns \edit{across settings}.\looseness=-1

Our study consisted of two main parts: a tutorial \edit{intended} to introduce the concepts and \edit{functionalities} needed to use our tool, followed by the main study in which participants answered questions about a \edit{pre-trained ML} model using the tool. We also conducted pre-study interviews to establish participants' background\edit{s} in ML and post-study interviews in which participants reflected on their experience\edit{s} with the tool and how they might use it in their ML pipeline\edit{s}. The study design was approved by our internal \edit{institutional review board}. \edit{Excerpts (screenshots) from the} Jupyter notebooks \edit{that we used in} the tutorial and \edit{in the} main study \edit{are in the appendix; the complete notebooks are available online.}\footnote{\url{http://github.com/dmelis/interpretwoe}}\looseness=-1

\subsubsection{Tutorial}\label{sec:tutorial}
\edit{\citet{kaur2020interpreting} found} that practitioners often use interpretability tools without fully understanding them, highlighting the importance of \edit{providing well-designed} tutorials and other accompanying documentation. For our user study, we designed a tutorial to introduce the \edit{concepts and functionalities needed to use our tool}, to evaluate \edit{participants'} understanding of these concepts, and to check whether \edit{their} responses in the main study were \edit{likely} based on a sound understanding\,---\,without being too time-consuming or tedious. After several iterations and pilot studies, we converged on an approximately 40-minute\edit{-long} tutorial based on a Jupyter notebook \edit{containing} equations, text, and images. This tutorial covered \edit{log-odds ratios}, weight of evidence, feature group for high-dimensional inputs, and sequential explanations for multi-class settings.\looseness=-1

\subsubsection{Main Study}
The goal of the main study was to assess participants' understanding of \edit{the WoE} and \edit{to investigate their} use of \edit{our interpretability} tool in the context of a realistic ML task. Participants were given a Jupyter notebook that included a dataset, an ML model trained using \edit{the} dataset, and our tool\edit{. They} were \edit{then} asked to answer several questions with the help of outputs (\edit{i.e.,} visualizations) from \edit{our} tool.\looseness=-1

The ML model was a random forest classifier trained \edit{using} the Online News popularity dataset~\citep{fernandes2015proactive}\edit{, which} consists of 39,797 news articles\edit{. Each article is} represented \edit{using} 59 features \edit{that} captur\edit{e} metadata about the article, \edit{such as its} length, \edit{any} links, and \edit{its sentiment} polarity. We trained the model to predict the category that the article was published under (\edit{e.g.,} ``Lifestyle'' or ``Business''), creating a six-class classification \edit{task}. We chose this dataset and \edit{this task} because the domain is understandable without expert knowledge or prior experience, the number of classes is large enough to \edit{permit} meaningful sequential explanations, and there are enough features to make explanations based on feature groups sufficiently different from \edit{explanations} based on individual features.

The main study was \edit{itself} divided into two parts, \edit{which were designed to let
us observe the use} of the two \edit{extensions} of \edit{the} WoE described in \edit{the
``Composite Hypotheses and Evidence'' section}: feature \edit{groups} and
sequential explanations. In the first part, participants were given
the option to view \edit{explanations} based \edit{on feature} groups \edit{or}
explanations \edit{based on individual features}, and were asked questions
that could be answered using either \edit{type} of \edit{explanation} (e.g., ``What
aspects of the news article contributed the most to this
prediction?''). \edit{This part of the study} was \edit{intended} to \edit{surface
participants'} preferences. In the second part, participants were given
the option to generate one-shot or sequential explanations, \edit{and} were
asked questions that could only be precisely answered using sequential
explanations (e.g., ``Why didn't the model predict [subset of
classes]?''). \edit{This part of the study} was \edit{intended} to \edit{assess} whether
participants could successfully use sequential
explanations.\looseness=-1

\subsubsection{Participants}
\edit{Potential} participants were recruited via email. To be considered for the study, they were asked to complete a survey about their \edit{ML background and their} experience with interpretability tools. Of 41 \edit{survey} respondents, we randomly selected 10 \edit{to participate in the study}. All \edit{participants} were ML practitioners (e.g., data scientists) with 1--20 years of experience. On average, \edit{participants} rated the role of ML in their jobs as 6.7 and their experience with interpretability tools as 3.2, both on a scale of 0 (``not at all'') \edit{to} 7 (``extremely''). \edit{P}articipants \edit{also rated} their familiarity with concepts from probability \edit{relevant to the WoE on a scale of 0 to 7}. Their average ratings were 2.7 for posterior \edit{class probabilities}, 3.3 for log likelihood\edit{s}, 3.2 for \edit{log-odds ratios}, and 0.9 for \edit{the WoE. On average, p}articipants took 1.7 hours to complete the study\edit{. Each participant was} compensated with a \$40 Amazon gift card.\looseness=-1

\subsubsection{Methods}
Participant\edit{s'} open-ended answers were scored by comparing them \edit{to} an answer key prepared in advance by two of the \edit{authors}. \edit{Answers that correctly identified key aspects (e.g., a feature with a large positive WoE value, pushing the model toward a particular prediction) were treated as correct even if the participants' specific language did not exactly match the language in the answer key.} To examine patterns of tool use, \edit{the} usability of \edit{our} tool, \edit{participants' interpretability} needs, and participants' general impressions, we analyzed automatically generated audio transcripts for high-level themes using inductive thematic analysis~\cite{braun2012thematic} and affinity diagramming. \looseness=-1

\subsection{Results}
\edit{T}he results \edit{from our} user study \edit{divide naturally} into three
categories: participants' \edit{understanding of relevant concepts}, tool
\edit{usability} and \edit{participants' preferences}, and \edit{general needs} for
interpretability tools. First, the pre-study interview\edit{s} and answers to
the checkpoint questions in the tutorial provide\edit{d} insight into
participants' understanding of concepts relevant \edit{to the WoE}. Second,
participants' approach\edit{es} to the questions \edit{in} the \edit{main study and} the\edit{ir}
\edit{patterns of tool use enabled us to examine} the usability of our
tool. Finally, \edit{via} the \edit{main} study and \edit{the} post-study interviews, we
\edit{were} able to \edit{uncover participants' general interpretability needs
and} additional criteria (\edit{beyond} our design principles) \edit{to consider
when designing and} evaluating interpretability tools.\looseness=-1

\subsubsection{Understanding of Relevant Concepts}
Our analysis \edit{showed}
that most  participants (7/10) struggled to understand and use prior class probabilities in the tutorial. The section on this topic was time-consuming: \edit{on average}, participants spent a third of their tutorial time on this section. Eventually, most participants either ignored the prior class probabilities or used them incorrectly, supporting the base rate fallacy. Nonetheless, participants were able to use \edit{the} WoE to \edit{correctly} answer questions in the main study for which prior class probabilities were relevant. This raises the possibility that \edit{although} they struggled with the abstract concept, they were able to use the information indirectly (e.g., via displayed class probabilities). This \edit{finding} is \edit{consistent} with \edit{those of \citet{kaur2020interpreting}}, who \edit{showed} that data scientists struggle to explain the \edit{concepts underlying} the explanations produced by \edit{generalized additive models~\cite{hastie1990generalized,Caruana2015-qf}} and SHAP \edit{\citep{lundberg2017unified}, even though they} still find these tools useful.\looseness=-1

Although participants generally understood the concept of WoE, some confused negative WoE values with negative values for input features, thus finding it challenging to make sense of the explanations. As a result, two participants provided incorrect answers for \edit{the} questions in the main study.

\subsubsection{Tool Usability and \edit{Participants' Preferences}}
Participants had no overwhelming preference between \edit{explanations based on feature groups and explanations based on individual features}. Indeed, they noted that the two levels of granularity provide complementary information, and switching between the two options was a clear pattern across all participants. \edit{Although} feature groups provide a high-level overview, making it ``easier to manage [reading the plot]...[and the] direction of analysis is a lot clearer'' (P8), \edit{explanations based on individual features} help participants in ``looking into more details in general...to know exactly which feature it was [that was responsible for a prediction]'' (P10).
We observed some differences in behavior based on participants' roles and expertise, though of course these are inconclusive with our small sample size.
Participants with more ML experience tended to rely on feature-level plots, while those with customer-facing jobs more often provided high-level answers based on feature groups, noting that feature~groups~``provide customer-friendly explanations'' (P8).\looseness=-1

Participants found sequential explanations to be a helpful breakdown of a larger explanation into parts. P3 noted these were like a ``story of how the prediction was made.'' Sequential explanations prompted more detailed answers to our questions and most participants (7/10) accurately answered questions in \edit{the second} part of the \edit{main} study using sequential explanations. They explained that the type of questions\,---\,which required understanding how each of the classes were ruled out\,---\,could not be answered via one-shot explanations. P8 commented, ``I find this to be quite helpful... I guess without this breaking it down to this point I wouldn't have thought twice really about this [input feature group] being a [differentiating] factor between the two [output classes]...I think that this would be a nice like final understanding [of the predicted class], this goes a lot deeper than I probably could have gone just looking at that without the tool. So I think it was very helpful in that case.''  \looseness=-1

Although most participants were happy with the level of detail presented, some participants with more ML experience expressed a desire for deeper understanding of how the explanations were generated. They understood the underlying concepts, but were wary of anything that appeared automated, including the breakdown of class comparisons in sequential explanations (which was automated) and the feature groups (which were actually manually generated).\looseness=-1

\edit{Even though} participants said that the tool helped them understand the model's predictions, not all of them envisioned the tool being added to their ML pipelines. Participants with significant prior ML experience already had established ways of ensuring that model predictions are reasonable, but
recognized other exciting use cases for the tool, such as communicating complex predictions to less experienced end users.
Particularly for high-risk domains, visualizations from the tool could help users probe odd predictions. P7 noted,
``With my focus on medical data, I do see the need in working with~a customer...there this [tool] would be a must-have. My team, we are engaged with customers and we have to educate them fast... So for me model interpretability there comes very close side by side with fairness.''\looseness=-1

\subsubsection{General Needs for Interpretability Tools}
Most interpretability tools, including ours, rely on tutorials \edit{and other accompanying} documentation to provide an introduction to the tool's concepts and functionalities. All participants appreciated the information presented in our tutorial: ``I can't imagine doing this [study] without the tutorial. I generally know a lot more about these concepts now'' (P5). The tutorial seemed to impact participants' overall accuracy in answering the \edit{questions in the} main study\,---\,those who spent longer on the tutorial tended to provide more accurate and more thoughtful answers. This manifested as longer time spent on exploring the tool in the tutorial and ensuring that their answers to the checkpoint questions were accurate and thorough. Even participants with less ML experience provided accurate answers when they devoted time to the tutorial. Participants appreciated the example in the tutorial and were able to generalize from \edit{this} example to \edit{the} questions the main study: ``The tutorial...helps you start in the right place. I went back to the example in the tutorial to~[determine how to] answer questions in the study'' (P6).\looseness=-1


Finally, participants expressed a desire to be able to more easily switch between different options (e.g., input features versus feature groups) rather than re-running code.
Interactivity was consequently the most commonly requested functionality in the post-study interview. This is in line with prior work on human-centered design principles for ML~\cite{amershi2019guidelines,hohman2019gamut,WB19}.\looseness=-1


\section{Limitations}

\edit{All} interpretability \edit{methods}, \edit{including ours}, involve various design
choices and assumptions (both implicit and explicit), many of which
give rise to potential limitations. \edit{First}, the \edit{concept} of
interpretability is notoriously ambiguous, and unlike \edit{supervised ML
tasks}, there is no ground truth to \edit{use for evaluation}, even for
proxy \edit{concepts like} feature importance.
\edit{As a result}, different \edit{interpretability methods} assume different
notions of interpretability, propose different quantities to
operationalize them, and (when needed) rely on different \edit{techniques} to
estimate the\edit{m. In turn,} these choices \edit{mean} that no interpretability
method \edit{will ever} be universally ideal. Moreover, summarizing the
behavior of complex models comes at a
price~\cite{rudin2018please}\,---\,that is, the explanations are partial,
only hold in a small neighborhood \citep{ribeiro2016why}, or make
strong assumptions about the data \citep{lundberg2017unified}. As a
result, explanations generated by one interpretability method \edit{seldom}
strictly \edit{dominate explanations generated by} another\edit{.} Furthermore,
different explanations might reveal information about different
aspects of the underlying model's behavior.\looseness=-1

Under this perspective, the design choices \edit{and assumptions involved in}
our interpretability method necessarily limit its scope and
applicability. Starting from our \edit{decision to distill characteristics}
of human explanation in\edit{to human-centered design principles, our method}
assum\edit{es} that \edit{human characteristics} are desirable for \edit{machine-generated
explanations. And, although the specific characteristics} that we \edit{focus
on yield a coherent} set of design principles, \edit{these principles are} not
exhaustive or universal. Some \edit{may} not be~necessary in all \edit{settings,}
and all are open to refinement.\looseness=-1

\edit{By relying on the concept of} weight of evidence \edit{(WoE) from information
theory, our} method inherits many of its strengths and
limitations. Concretely, there are three main settings in which there
is a clear case for using explanations based on the WoE: \edit{1)} when the
underlying model is generative, \edit{2) when the underlying model is} log
linear, and \edit{3) when the underlying model is} a multi-class classifier. We provide a detailed discussion of these three settings in the appendix.
In terms of limitations, the WoE requires access to \edit{the} conditional
likelihoods \edit{$P(X_{\mathcal{A}_i} | X_{\mathcal{A}_{i-1}}, \dots,
X_{\mathcal{A}_1}, Y)$}, which limits its use to settings in which
these are accessible or can be accurately estimated from finite
samples\edit{. Estimating densities for more complex types of data, such as
images,} is an active area of research, and \edit{although it may be possible
to integrate} new advances \edit{into our method, its} applicability
to \edit{such types of data} is \edit{currently} limited. Other important design
choices \edit{involved in our method} include the \edit{technique} for partitioning
\edit{the} classes \edit{in multi-class settings to yield} sequential explanations
and the \edit{technique for partitioning input} features into feature
\edit{groups}. Although we \edit{chose} generic solutions to these \edit{challenges, there
are other techniques, which may be more appropriate in some settings,}
which \edit{will} invariably lead to different explanations. We \edit{defer a}
thorough \edit{investigation} of \edit{these choices} for future work.\looseness=-1

The main limitations of our user study \edit{are the} number of participants,
\edit{the type of participants,} and the extent to which the study conditions
\edit{mimic a realistic setting}. We \edit{chose to conduct an} artifact-based
interview \edit{study to ensure sufficient interaction time and nuanced data
collection for each participant, but this limited} the number of
participants \edit{that we could consider, thereby precluding} statistical
analys\edit{e}s. Our participants were also limited to \edit{ML}
practitioners. \edit{Following previous}
work \cite[e.g.,][]{kaur2020interpreting}, we \edit{chose to focus on ML
practitioners} because \edit{they} are frequent users of interpretability
tools in the wild. \edit{Finally, although} we tried to \edit{design the study so
as to mimic} a realistic setting, we cannot be sure that this
experience was representative of participants' day-to-day \edit{experiences}
(e.g., working with their own datasets). Ideally, \edit{we would have} run a
longitudinal field study with multiple types \edit{of participants to enable
us to} observ\edit{e participants'} tool use over time as \edit{they} gain expertise
in using it. However, this would \edit{have} require\edit{d} additional resources
(e.g., to support multiple types of data) \edit{and} was \edit{therefore
infeasible}. Instead, our \edit{user study} serves as an initial evaluation \edit{of
our} interpetability \edit{method}.\looseness=-1

\section{Discussion}

\edit{In this paper, we take inspiration from the study of human explanation,
drawing on the literature on human explanation in philosophy,
cognitive science, and the social sciences to propose a list of design
principles for machine-generated explanations that are meaningful to
humans. We develop a method for generating explanations that adhere to
these principles using the concept of weight of evidence from
information theory. We show that this method can be adapted to meet
the needs of modern ML\,---\,that is, high-dimensional, multi-class
settings\,---\,and that the explanations can be estimated accurately from
finite samples, are robust to small perturbations~of the inputs, and
are usable by ML practitioners.}\looseness=-1

\edit{This paper opens several avenues for future work. Adapting modern
density estimation methods for complex types of data, such as images,
might hold the key to wider applicability of our method. Regarding
evaluation, an immediate next step would be to carry out a follow-up
user study to investigate various design choices, such as the
technique for partitioning the classes in multi-class
settings. Ideally, future user studies should involve participants'
own models and should rely on questions that attempt to uncover
insights that are relevant to their day-to-day
experiences.}\looseness=-1

\edit{The findings from our user study offer important lessons that we
believe are generally applicable to other interpretability
tools. Chief among these is the importance of user-friendly and
engaging tutorials that provide users with the necessary understanding
of the tool and its intended usage, and users' desire for flexibility
in tools. These results underscore the importance of putting human
needs at the center of the design and evaluation of interpretability
methods. The human computation community is uniquely situated to drive
this work as it requires interdisciplinary expertise in both ML and
HCI, fields that are central to HCOMP.}\looseness=-1

\edit{In the spirit of developing AI responsibly, we believe that papers
proposing new interpretability methods should also provide a
discussion, as we have done in the previous section, of not only those
settings for which the proposed method is suitable, but also those
settings that fall outside its scope. In addition, we recommend that
authors should explicitly describe the notion of interpretability (or
explanation) that they aim to operationalize, allowing readers to
situate the contributions in relation to other interpretability~methods and to understand their scope and applicability.}\looseness=-1

\bibliography{references.bib}

\clearpage
\pagebreak

\appendix

\section{Design Principles in Further Detail}\label{sec:principles_detail}

\edit{W}e surveyed the literature on the nature of explanation from
philosophy, cognitive science, and the social sciences, leading us to
propose a list of design principles for machine-generated explanations
that are meaningful to humans. In~this section, we further discuss
\edit{each of} these principles.\looseness=-1

\paragraph{D1. Explanations should be contrastive.} A recurring theme across disciplines is the hypothesis that humans tend to explain in contrastive terms. We therefore argue that machine-generated explanations should also be contrastive. In other words, the \emph{explanandum} should be the question of why a model $\cal{M}$ predicted a particular output $y$ for some input $x$ \emph{instead of predicting} $y'$. However, most current interpretability methods do not explicitly refer to an alternative~$y'$, implicitly assuming it to be the complement of $y$.\looseness=-1

\paragraph{D2. Explanations should be exhaustive.} The explanations produced by most current interpretability methods refer only to why the input $x$ points to a single hypothesis (i.e., the prediction $y$) rather than ruling out all alternatives. However, human explanations tend to provide a justification for why every alternative $y'$ was not predicted. This is important for (at least) two reasons: First, it is necessary for generating explanations that are counterfactual. Second, it entails explanations that are, to use Hempel's terminology, \textit{complete}.\footnote{Not to be confused with  \textit{complete} according to the terminology~of \citet{goodman2006intuitive}, where all variables are explained.} For example, an explanation for a diagnosis of pneumonia based
solely on the presence of a cough is not exhaustive because a cough is
a symptom of other diseases too.\looseness=-1

\paragraph{D3. Explanations should be modular and compositional.} Explanations are most needed for the predictions of complex black-box models like neural networks. Yet, when used in these settings, most current interpretability methods generate a single, high-dimensional static explanation for each prediction (e.g., a heatmap for a prediction made by an image classifier). Explanations like these can be difficult to understand and draw conclusions from, particularly for
non-expert users. In addition, this approach differs from human
explanations, which tend to decompose into simple
components~\citep{hempel1962deductive}. We therefore argue that
machine-generated explanations should also explain using multiple
simple accumulative statements, each addressing a few aspects of the
evidence. We note, however, that this type of modularity introduces an inherent trade-off between the number of components in an explanation (too many
components might be difficult to understand simultaneously) and their
relative complexity (simpler components are easier to draw conclusions
from, but using simpler components may mean that more of them are
needed). Breaking up predictions into simple components also helps shift
interpretability methods from treating explanations as a product
towards explanations as a process~\citep{lombrozo2012explanation,miller2019explanation}.\looseness=-1

\paragraph{D4. Explanations should rely on easily-understandable quantities.} Explanations are only as useful as the information that they provide to users. We therefore argue that explanations should rely on quantities that can be easily understood, thereby making it less likely that users will misunderstand them, potentially leading to misuse~\citep{kaur2020interpreting}.\looseness=-1

\paragraph{D5. Explanations should be parsimonious.} As noted by \citet{miller2019explanation}, one of the most salient recurring themes in the literature on human explanation is minimality\,---\,i.e., only the most relevant facts should be included in explanations. We therefore argue that explanations should adhere to Occam's Razor, which states that the simpler of two equally good explanations should be preferred. Furthermore, if omitting less-relevant facts makes a machine-generated explanation easier to understand while remaining equally faithful~to~the underlying model, then they should be omitted.\looseness=-1

\section{Axiomatic Derivation of the WoE}\label{sec:axiomatic}

\edit{Interpretability} is fundamentally a human-centered concept. As such,
human needs should therefore be at the center of both the design and
evaluation of interpretability methods. We achieve this by first
proposing a list of design principles that we argue human-centered
machine-generated explanations should satisfy and then realizing these
design principles starting from the concept of weight of evidence
(WoE)\edit{.}

\citet{good1985weight} provides an axiomatic derivation for the WoE of $e$ in favor of $h$, as defined in Equation~\eqref{eq:woe_def}. This derivation shows that the WoE is (up to a constant) the only function $F$ of $e$ and $h$ that satisfies the following properties:\looseness=-1
\begin{enumerate}
    \item $F(e,h)$ depends only on $P(e\mid h)$ and
    $P(e\mid \overline{h})$.  \item $P(h\mid e)$ is a function of only
    $P(h)$ and $F(e,h)$.  \item When $e$ is decomposable, $F(e, h)$ is
    additive in terms of $e$, e.g., $F(e_1 \land e_2, h) = F(e_1, h) +
    F(e_2,h)$.
\end{enumerate}

We note that all three of these properties are fairly general. For example, the second property is a reasonable requirement for any probabilistic model of feature importance. Meanwhile, the first property is a sufficient (although not a necessary) condition for obtaining contrastive explanations.

\section{Well-Known Model Families}\label{sec:instantiation}

In this section, we show that for well-known families of simple
classifiers, the WoE has a closed-form expression. Consider a binary
classification setting\,---\,i.e., $y\in\{0,1\}$\,---\,with $n$ input
features\,---\,i.e., $x=(x_1,\dots,x_n)$\,---\,and, for~conciseness, let
$p_1=P(Y=1)$ and $p_0=P(Y=0)$.\looseness=-1

\subsection{Logistic Regression}
For logistic regression, the predictive posterior is
$$P(Y=1 \mid X) = \sigma(w^\top x+w_0) \quad \in [0,1],$$ where
$\sigma(t) = \frac{e^t}{e^t+1}= \frac{1}{1+e^{-t}}$ is the sigmoid
function. Logistic regression already has a natural log-odds
interpretation:
\begin{align*}
    \log\frac{P(Y=1\mid X)}{P(Y=0\mid X)} &= \frac{\sigma(w^\top x+w_0)}{1-\sigma(w^\top x+w_0)} \\ &= \log e^{w^\top x+w_0} = w^\top x +w_0 .
\end{align*}
From this, the WoE is easy to compute and unsurprising:
\begin{align*}
    \text{woe}(Y=1 : X=x) &= \log\tfrac{P(Y=1\mid X)}{P(Y=0\mid X)}/\log\tfrac{P(Y=1)}{P(Y=0)} \\ &= w^\top x + w_0  - \log\tfrac{p_1}{p_0} .
\end{align*}
Therefore, the WoE for logistic regression can be interpreted as removing the effect of the base rate (via the log-odds ratio $\log\tfrac{p_1}{p_0}$) from the linear attribution model (i.e.,  $w^\top x + w_0$).

Unfortunately, obtaining closed-form expressions for per-feature WoE
values is not as simple. Because logistic regression is not a
generative model, it does not explicitly model the conditional
probabilities needed to obtain these values. For example, neither
$P(Y=1 \mid X_i)$ nor $P(X_i \mid Y=1)$ can be unambiguously obtained
from $P(Y=1 \mid X)$. To determine these conditional probabilities,
assumptions must be made. For example, the simplest coherent
identification of the coefficients with per-feature log-likelihood
ratios is:
\begin{align*}
	w_0 &= \log\frac{p_1}{p_0} \\
	w_ix_i + w_0 &= \log\frac{P(Y=1\mid x_i)}{P(Y=0\mid x_i)} .
\end{align*}
However, this identification makes the assumption that the coefficients do not model feature interactions. Under this~identification, the WoE value for $X_i$ is then as follows:
\begin{equation*}
	\text{woe}(Y=1: X_i) = w_ix_i .
\end{equation*}
This is a natural notion of feature importance for log-linear models under a conditional independence assumption. The posterior log odds of $Y = 1$ versus $Y = 0$ is then given by:
\begin{align*}
  &\log\tfrac{P(Y = 1 \mid X)}{P(Y =0 \mid X)} \notag\\
  &\quad =\sum_{i=1}^n\text{woe}(Y\!=\!1\!:\!x_i) + \log\frac{p_1}{p_0}= w^\top x +w_0 .
\end{align*}

\edit{Other interpretability} methods can be shown to recover logistic regression's coefficients too. For example, it is easy to show that without penalization, with a logistic link function, and asymptotically in the number of fitting samples, LIME \citep{ribeiro2016why} would find a surrogate model that coincides with the true log-linear model, so the resulting explanations would coincide with the WoE values.\looseness=-1

\subsection{Na\"ive Bayes}

The equivalence between the WoE and na\"ive Bayes (NB) is well known,
but we revisit it here for completeness. An NB~classifier already
assumes conditional independence:
\[ P(X_i \mid Y) \perp P(X_j \mid Y) \quad \forall i\neq j. \]
The conditional probabilities are modeled as
\[ p(Y=1 | X)=P(Y=1)\prod_{i=1}^nP(X_i \mid Y=1), \]
where $P(X_i \mid Y)$ is estimated during training. Therefore,~for a NB classifier, the total WoE is defined as follows:
\begin{align*}
\text{woe}(Y=1 : X) &=  \log\frac{P(X\mid Y=1)}{P(X\mid Y=0)} \\
    &= \log\frac{\prod_{i=1}^nP(X_i \mid Y=1)}{\prod_{i=1}^nP(X_i \mid Y=1)} \\&= \sum_{i=1}^n \log \tfrac{P(X_i \mid Y=1)}{P(X_i \mid Y=0)}.
\end{align*}
In addition,
\begin{align*}
    \text{woe}(Y=1 : X_i) &= \text{woe}(Y=1 : X_i | X_0,\dots,X_{i-1})  \\&= \log \tfrac{P(X_i \mid Y=1)}{P(X_i \mid Y=0)} .
\end{align*}

A Gaussian \edit{NB} (GNB) classifier would \edit{parametrize} the~conditional likelihoods as $n$-dimensional Gaussians:
\begin{multline*}
    p(X \mid Y=k) = \\  \frac{1}{\sqrt{(2\pi)^n|\Sigma_k|}} \exp \left \{ -\frac{1}{2}(x-\mu_k)^\top \Sigma_i^{-1}(x-\mu_k), \right \}
\end{multline*}
where the covariance matrix $\Sigma_k$ is diagonal in order to satisfy the NB assumption. This then implies the following:
\begin{align*}
  &p(X_i \mid Y\!=\!k) \notag\\
  &\quad = \frac{1}{\sqrt{(2\pi)^n[\Sigma_k]_{ii}}} \exp \left \{ -\frac{1}{2[\Sigma_k]_{ii}}|x_i-\mu_{ki}|^2 \right \}.
  \end{align*}
Therefore,
\begin{align*}
  &\text{woe}(Y=1 : X_i) \notag\\
  &\quad= \log \left( \frac{[\Sigma_0]_{ii}}{[\Sigma_1]_{ii}}\right)^{1/2} \exp \left \{ -\frac{1}{2} \left( \tfrac{1}{[\Sigma_1]_{ii}}|x_i - \mu_{1,i}|^2 \right.\right.\\
  &\quad\quad-\left.\left. \tfrac{1}{[\Sigma_0]_{ii}}|x_i - \mu_{0,i}|^2  \right) \right \} \notag \\
  &\quad= -\frac{1}{2} \left( \tfrac{1}{[\Sigma_1]_{ii}}|x_i\!-\!\mu_{1,i}|^2  - \tfrac{1}{[\Sigma_0]_{ii}}|x_i\!-\!\mu_{0,i}|^2  \right)\notag\\
  &\quad\quad{}-\frac{1}{2}\log\frac{[\Sigma_1]_{ii}}{[\Sigma_0]_{ii}}
\end{align*}
and
\begin{align*}
  &\text{woe}(Y=1 : X) \notag\\
  &\quad=-\frac{1}{2} \biggl(x^\top(\Sigma_1^{-1} - \Sigma_0^{-1})x -2(\mu_1^\top\Sigma_1^{-1} - \mu_0^\top \Sigma_0^{-1})x \\
	&\quad\quad+ (\mu_1^\top\Sigma_1^{-1}\mu_1 - \mu_0^\top\Sigma_0^{-1}\mu_0 ) \biggr )
	- \frac{1}{2}\log\frac{|\Sigma_1|}{|\Sigma_0|} .
\end{align*}
This implies that $\text{woe}(Y=1 : X_i) >0$ if and only if:
\[ \frac{1}{\Sigma_{1,ii}} |x_i - \mu_{1,i}|^2  - \log\Sigma_{1,ii} < \frac{1}{\Sigma_{0,ii}} |x_i - \mu_{0,i}|^2 -\log \Sigma_{0,ii}. \]

In other words, $\text{woe}(Y=1 : X_i) >0$ if and only if $x_i$ is closer (according to a variance-normalized distance) to  $\mu_{1,i}$ than to $\mu_{0,i}$. Similarly, the total WoE satisfies the following:\looseness=-1
\begin{align*}
  &\text{woe}(Y\!=\!1\!\mid\!X)>0 \Longleftrightarrow \notag\\
  &\quad d^2_{\Sigma_1}(x,\mu_1) - d^2_{\Sigma_0}(x,\mu_0) < \log\tfrac{|\Sigma_1|}{|\Sigma_0|},
\end{align*}
where $d_\Sigma$ is the Mahalanobis distance with covariance $\Sigma$.\looseness=-1

\subsection{LDA and QDA}

Linear discriminant analysis (LDA) and quadratic discriminant analysis (QDA) are similar to Gaussian na\"{i}ve Bayes, except that they do not make a conditional independence assumption\,---\,i.e., the covariance matrix need not be diagonal. Although non-conditional per-feature WoE values can still be computed easily (because marginalizing a multivariate Gaussian boils down to taking subsets of the mean and covariance matrix), the conditional per-feature WoE values do not have as simple an expression. In the general case, where~the variables are not independent, if $X \mid Y \sim \mathcal{N}(\mu, \Sigma)$, then $X_i \mid X_1,\dots,X_{i-1},Y \sim \mathcal{N}(\tilde{\mu}, \tilde{\sigma})$, where\looseness=-1
\begin{align*}
	\tilde{\mu} &= \mu_i + \Sigma_{[i,1:i-1]}\Sigma_{[1:i-1,1:i-1]}^{-1}(x_{[1:i-1]} - \mu_{[1:i-1]})\\
	\tilde{\sigma} &= \Sigma_{ii} - \Sigma_{[i,1:i-1]}\Sigma_{[1:i-1,1:i-1]}^{-1}\Sigma_{[1:i-1,i]} .
\end{align*}
Setting off-diagonal elements to be $0$, we recover $$P(X_i \mid
X_{i-1},\dots,X_{1},Y)\sim\mathcal{N}(\mu_i, \Sigma_{ii}).$$ We can use the values of $\tilde{\mu}$ and $\tilde{\sigma}$
above to derive closed-form (albeit cumbersome) expressions for
$\text{woe}(Y=1 : X_i \mid X_{i-1},\dots,X_1)$ in terms of log-density
ratios of normal distributions. The total WoE is analogous to the total WoE  for GNB. For example, LDA makes a
homoscedasticity assumption ( $\Sigma_1=\Sigma_0=\Sigma)$, so the
total WoE satisfies
\[ \text{woe}(Y=1 \mid X)>0 \Longleftrightarrow d^2_{\Sigma}(x,\mu_1) - d^2_{\Sigma}(x,\mu_0) < 0 . \]

\section{\edit{Use Cases and Limitations}}\label{sec:limitations}

In this section, we discuss how \edit{best} to choose between \edit{our} method and
other methods\edit{, such as LIME \citep{ribeiro2016why} and
SHAP \citep{lundberg2017unified}}.\looseness=-1

\subsection{No Explanation is Inherently ``Better''}\label{sec:no_better_explanation}

In the \edit{``}Design Principles\edit{'' section,} we proposed a list of design principles \edit{for machine-generated explanations,} distilled from the literature on the nature of explanation from philosophy, cognitive science, and the social sciences. However, distilling universal principles for what makes one explanation better than another is much more difficult. Indeed, the answer to this question is highly context dependent. This implies that explanations generated by different interpretability methods might be useful in different situations, depending on the type of model, the data, the user, and the intended~use~of the explanations~\cite{VW21}.\looseness=-1

Further complicating matters is the fact that post-hoc interpretability methods are, by definition, imperfect. Summarizing the behavior of complex models comes at a price~\cite{rudin2018please}\,---\,that is, the explanations are partial, only hold in a small neighborhood \citep{ribeiro2016why}\edit{,} or make strong assumptions about the data \citep{lundberg2017unified}. As a result\edit{,} explanations generated by one interpretability method \edit{seldom} strictly dominate explanations generated by another: there are always scenarios in which one method's assumptions and approximations will be more \edit{appropriate} than \edit{those of another}. Furthermore, different explanations might reveal information about~different aspects of the underlying model's behavior.\looseness=-1

Based on this premise, we next discuss when WoE explanations are likely to be useful\,---\,as complements to or substitutes for \edit{other} interpretability methods\,---\,and when they are not. Our goal is to provide guidelines for deciding when to use our method and when to use other methods. We focus on LIME \citep{ribeiro2016why} and SHAP \citep{lundberg2017unified} because \edit{they} are very widely used.\looseness=-1

\subsection{When Are WoE Explanations Useful?}\label{sec:woe_use_cases}

We first discuss three main settings in which there is a clear case
for using explanations based on \edit{the} weight of evidence.

\vspace{0.05cm}
\noindent
\textit{The underlying model is generative.} Explaining the predictions made by a generative model in generative terms (see the \edit{``Relation to LIME and SHAP'' section}) seems, conceptually, a better fit than doing so in discriminative terms. But there are other tangible reasons to use WoE explanations for generative models too: First, generative models already compute conditional likelihoods \edit{internally. Even when} these conditional likelihoods are not accessible (e.g., as is the case with black-box models), it is reasonable to think that they will be stable, as will be their empirical estimates, even for small sample sizes. Indeed, we showed in the \edit{``}Quantitative Experiment\edit{s''} section that when the underlying model is generative, WoE values can be very accurately estimated.\looseness=-1

\vspace{0.05cm}
\noindent
\textit{The underlying model is log linear.} If the underlying model is log linear or yields simple log-odds ratios, then the WoE might have a closed-form, or at least simple, expression. For example, for logistic regression, the log odds are linear, \edit{as we explained above}. This differs from post-hoc interpretability methods like LIME and SHAP, which need to be modified to accommodate non-linear link functions.\looseness=-1

\vspace{0.05cm}
\noindent
\textit{The underlying model is a multi-class classifier.}  In their original forms, both LIME and SHAP produce one-shot explanations in for the predicted class. When the number of classes is large, this is unlikely to yield useful explanations. The WoE, on the other hand, naturally extends to settings where the hypotheses $h$ and $h'$ to contrast correspond to sets of classes, as discussed in the main text.
Indeed, in our user study, participants found that sequential
explanations helped them to understand how each of the classes were
ruled out.

Even when the underlying model is not generative or log linear, it might still be desirable to explain its predictions in generative terms, perhaps as a complement to other post-hoc interpretability methods. However, WoE explanations should only be used if the following \edit{two} conditions hold:\looseness=-1

\vspace{0.05cm}
\noindent
\textit{Conditional likelihoods are accessible or can be accurately estimated from finite samples.} \edit{The} meta-algorithm for generating WoE explanations requires access to the conditional likelihoods $P(X_{\mathcal{A}_i} | X_{\mathcal{A}_{i-1}}, \dots, X_{\mathcal{A}_1}, Y)$. If the \edit{underlying model computes these} conditional likelihoods internally, then they can be used directly; however, in other cases, they must be estimated from samples. Our \edit{quantitative experiments} suggest that the WoE can be accurately estimated\,---\,even in relatively high dimensions\,---\,from finite samples.\looseness=-1

\vspace{0.05cm}
\noindent
\textit{Log-odds ratios are \edit{easily understandable by} the stakeholders \edit{who will use} the explanations.} Following Principle 4, explanations should rely on easily understandable quantities. As we discussed in the ``Weight of Evidence'' section, there is ample evidence to suggest that the units in which the WoE is expressed (log-odds ratios) are easily understandable. However, we recommend experimentally verifying that this is indeed the case in the context in which WoE explanations~will be used, for the stakeholders who will use them. \looseness=-1

\section{\edit{Sequential Explanations}}\label{sec:implementation_challenges}

\begin{figure}
    \centering \includegraphics[scale=0.18, trim={0 2.2cm 18cm 0},
    clip]{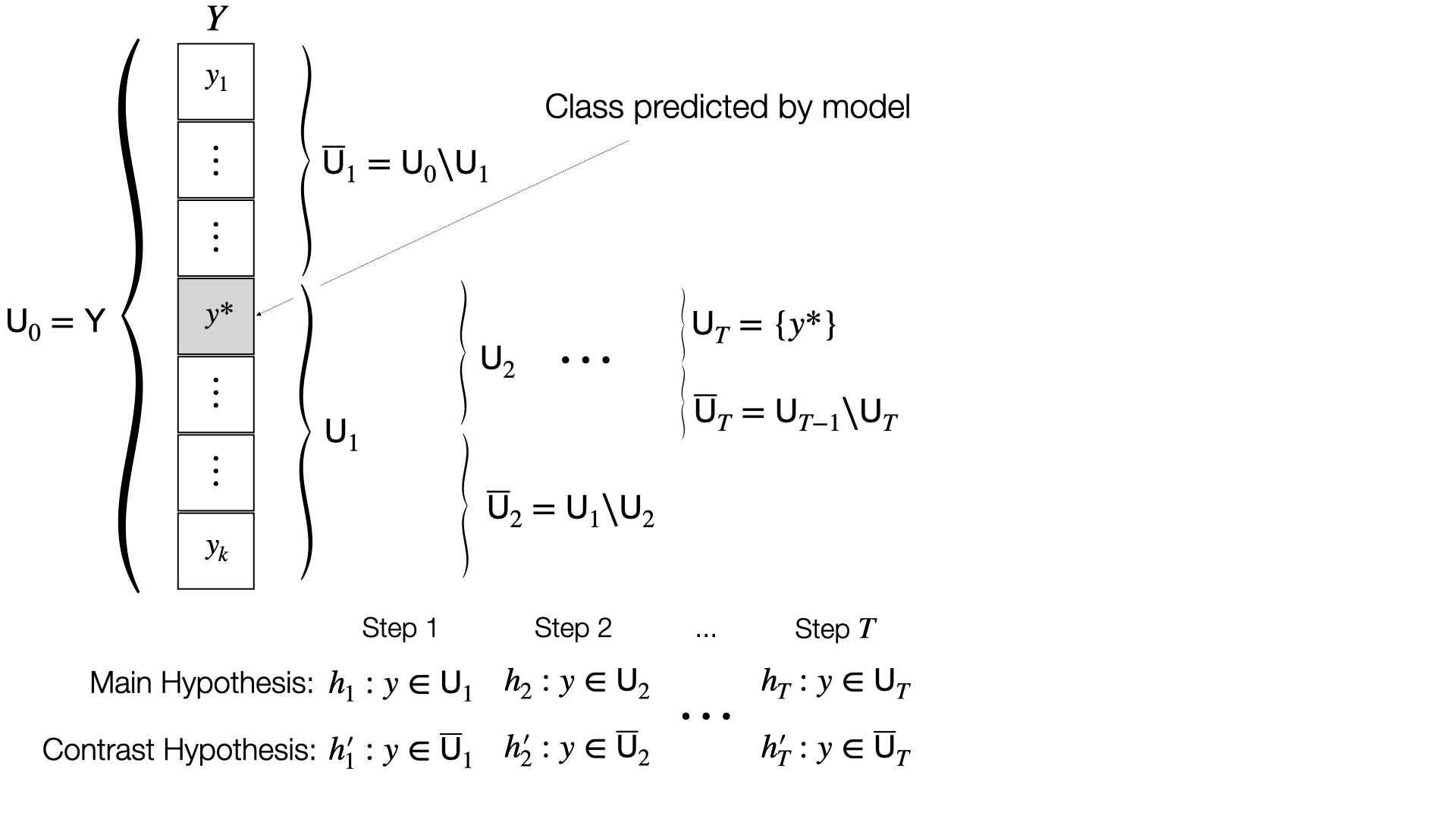} \caption{\edit{Our} method breaks up
    \edit{up} explanations \edit{in multi-class settings} into sequences
    of ``smaller'' explanations.}  \label{fig:nested}
\end{figure}

Figure~\ref{fig:nested} provides a schematic illustration of the procedure for generating sequential explanations that our tool uses in settings where the number of classes is very large. At each step, a subset of the possible classes $\mathsf{U}_i$ is ruled out. By design,
the subset that is retained contains the predicted class $y^*$. In
other words, with each subsequent step, a smaller set of classes is
\edit{retained} until $h$ consists of only $y^*$. At the end of this sequential \edit{procedure}, every~alternative $y'$ \edit{will have} been ruled out (c.f.~Principle 2).\looseness=-1

\section{Jupyter Notebooks}\label{sec:notebooks}

\edit{In the rest of the appendix}, we provide excerpts from the Jupyter
notebooks that we used in the tutorial and in the main study. The
complete notebooks are \edit{available
online}.\footnote{\edit{\url{http://github.com/dmelis/interpretwoe}}}\looseness=-1

\begin{figure*}[!ht]
    \centering
    \includegraphics[width=0.475\linewidth, clip, trim={1.8cm 2.5cm 2cm 2cm}, page=1]{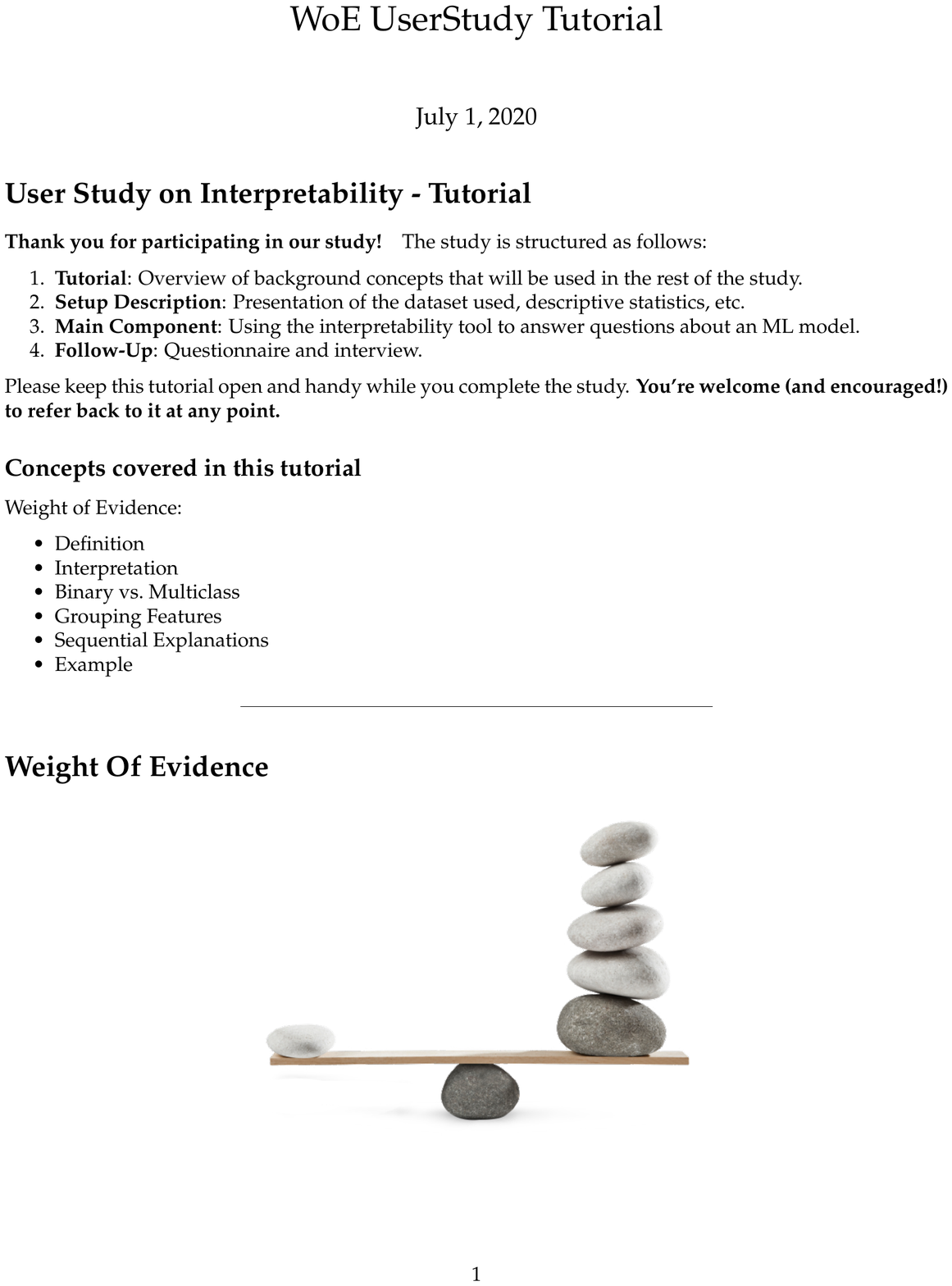}%
    \includegraphics[width=0.475\linewidth, clip, trim={1.8cm 2.5cm 2cm 2cm}, page=2]{figs/WoE_UserStudy_Tutorial.pdf}
    \includegraphics[width=0.475\linewidth, clip, trim={1.8cm 3cm 2cm 2cm}, page=3]{figs/WoE_UserStudy_Tutorial.pdf}%
    \includegraphics[width=0.475\linewidth, clip, trim={1.8cm 3cm 2cm 2cm}, page=4]{figs/WoE_UserStudy_Tutorial.pdf}
    \caption{Jupyter notebook for the tutorial.}
    \label{fig:tutorial_1}
\end{figure*}

\clearpage
\pagebreak

\begin{figure*}[!ht]
    \centering
    \includegraphics[width=0.475\linewidth, clip, trim={1.8cm 2.5cm 2cm 2cm}, page=5]{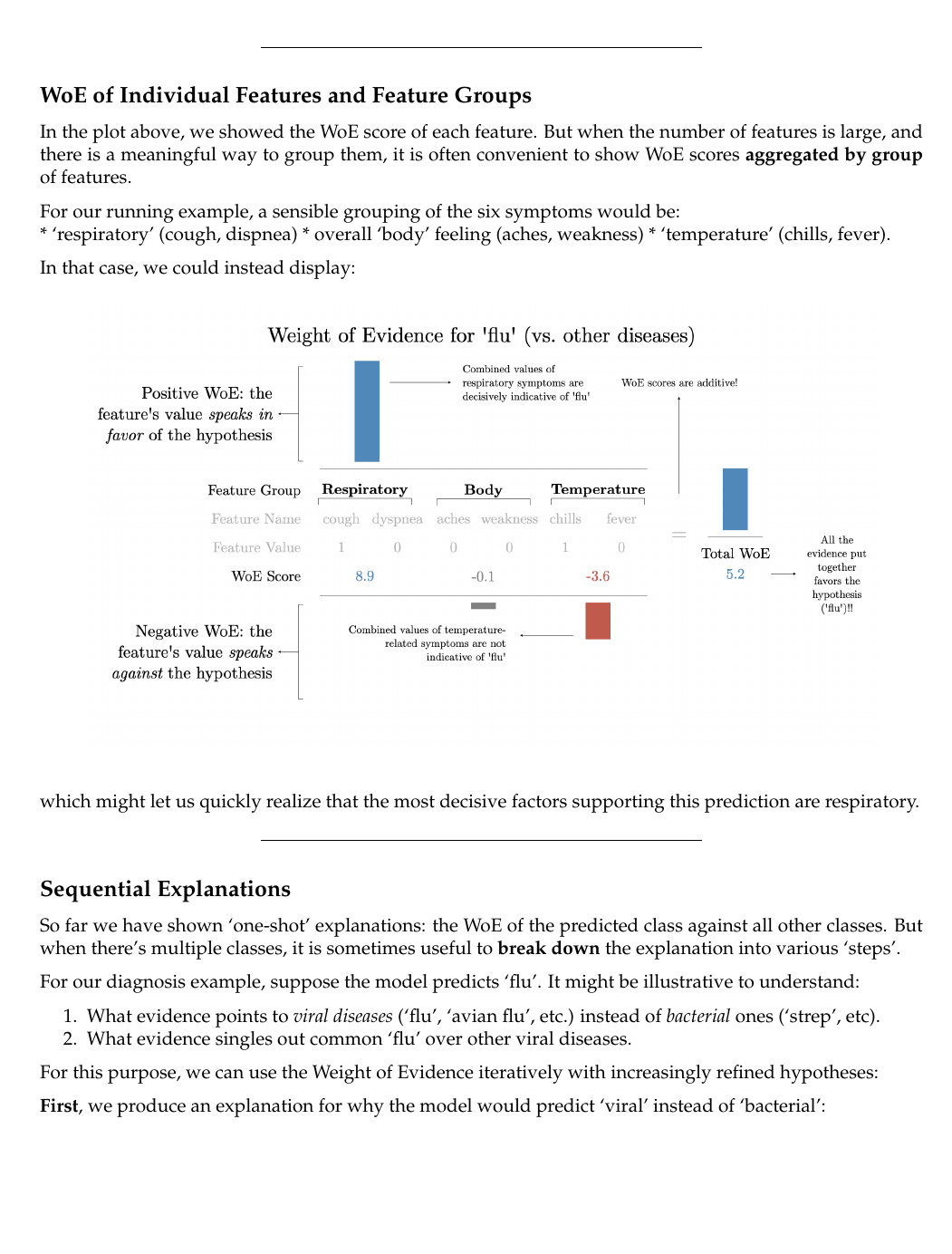}%
    \includegraphics[width=0.475\linewidth, clip, trim={1.8cm 2.5cm 2cm 2cm}, page=6]{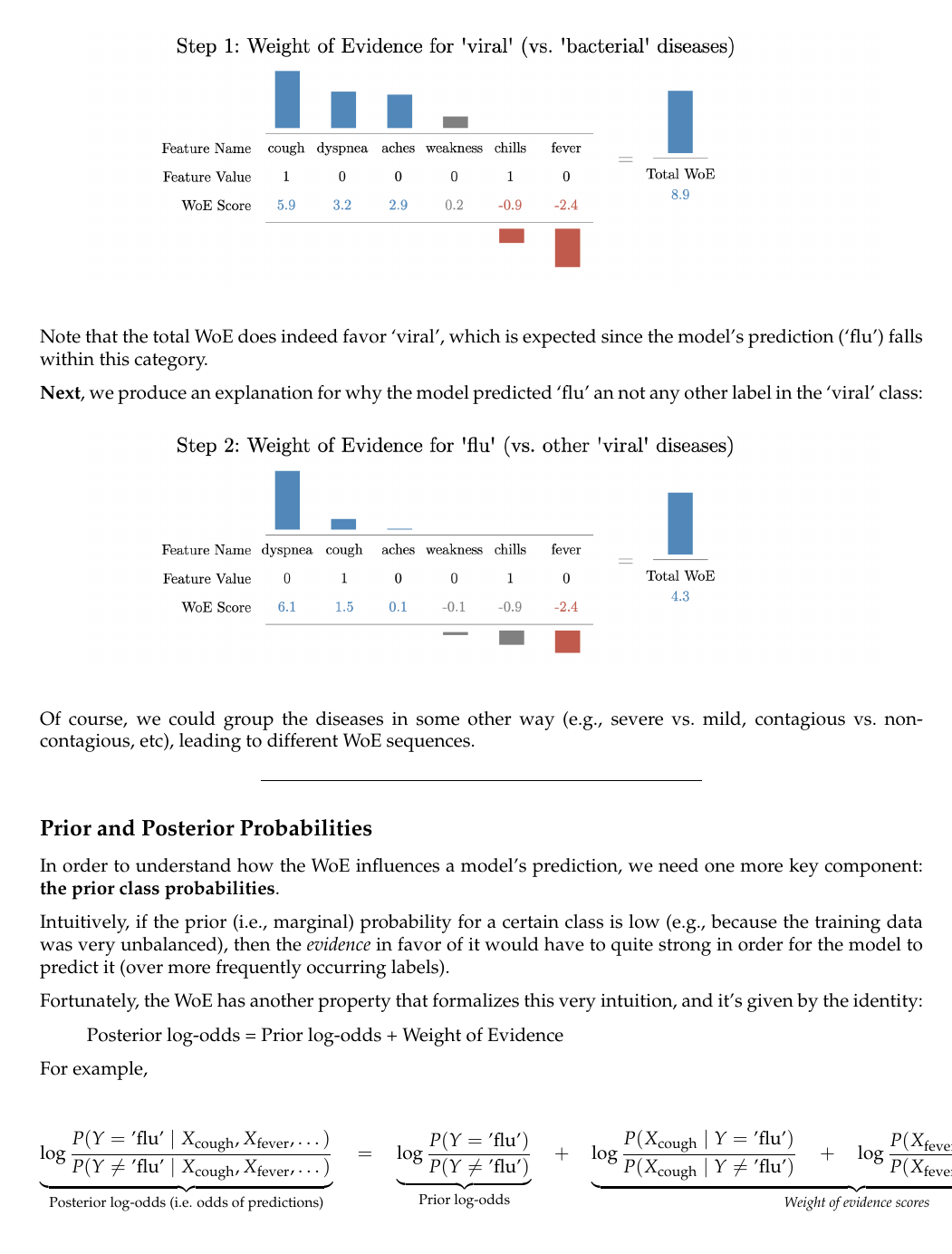}
    \includegraphics[width=0.475\linewidth, clip, trim={1.8cm 3cm 2cm 2cm}, page=7]{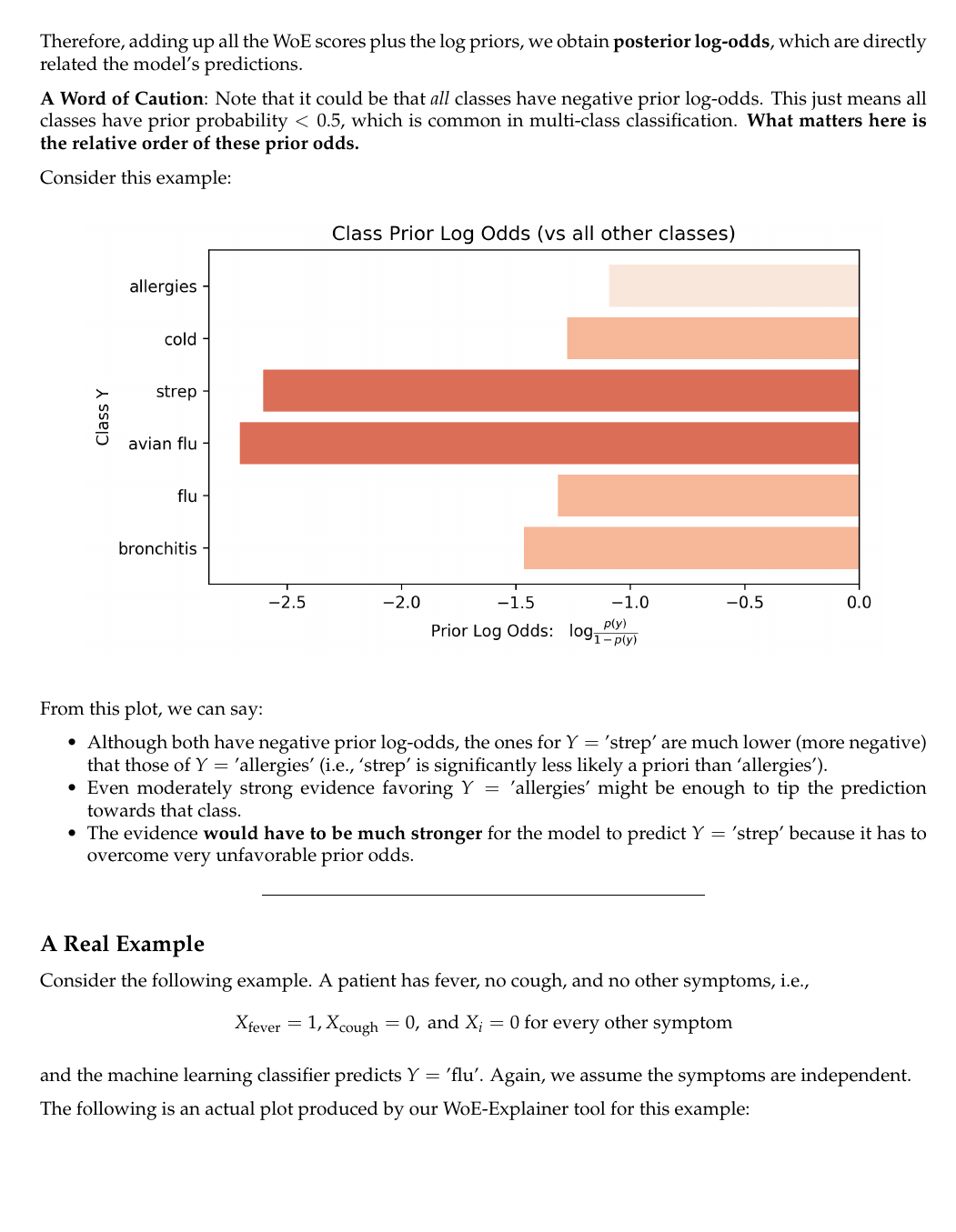}%
    \includegraphics[width=0.475\linewidth, clip, trim={1.8cm 3cm 2cm 2cm}, page=8]{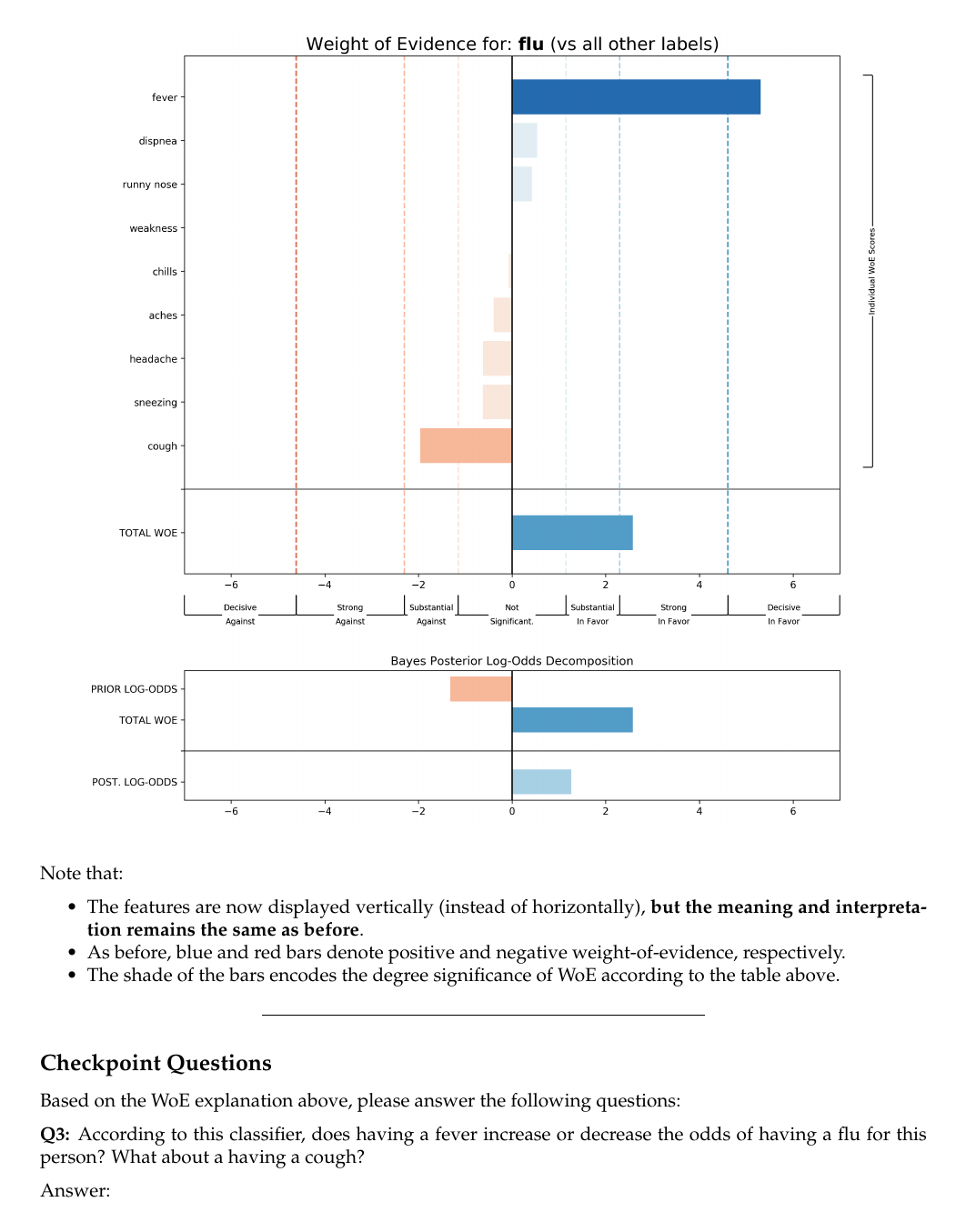}
    \caption{Jupyter notebook for the tutorial (continued).}
    \label{fig:tutorial_2}
\end{figure*}

\clearpage
\pagebreak

\begin{figure*}[!ht]
    \centering
    \includegraphics[width=0.475\linewidth, clip, trim={1.8cm 2.5cm 2cm 2cm}, page=1]{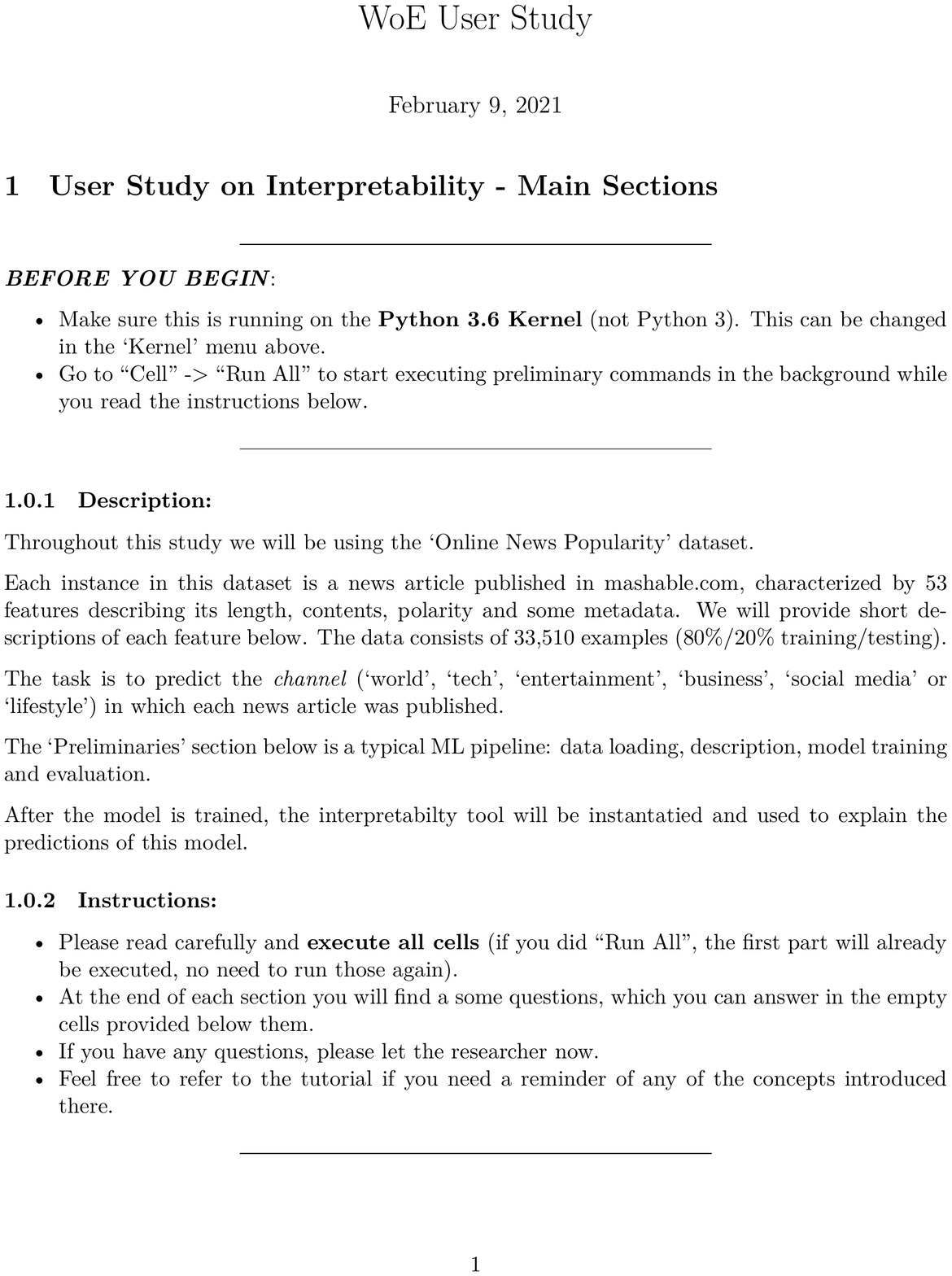}%
    \includegraphics[width=0.475\linewidth, clip, trim={1.8cm 2.5cm 2cm 2cm}, page=2]{figs/WoE_UserStudy_Main.pdf}
    \includegraphics[width=0.475\linewidth, clip, trim={1.8cm 3cm 2cm 2cm}, page=5]{figs/WoE_UserStudy_Main.pdf}%
    \includegraphics[width=0.475\linewidth, clip, trim={1.8cm 3cm 2cm 2cm}, page=6]{figs/WoE_UserStudy_Main.pdf}
    \caption{Jupyter notebook for the main study (some output pages omitted for brevity).}
    \label{fig:study_1}
\end{figure*}

\clearpage
\pagebreak

\begin{figure*}[!ht]
    \centering
    \includegraphics[width=0.475\linewidth, clip, trim={1.8cm 2.5cm 2cm 2cm}, page=8]{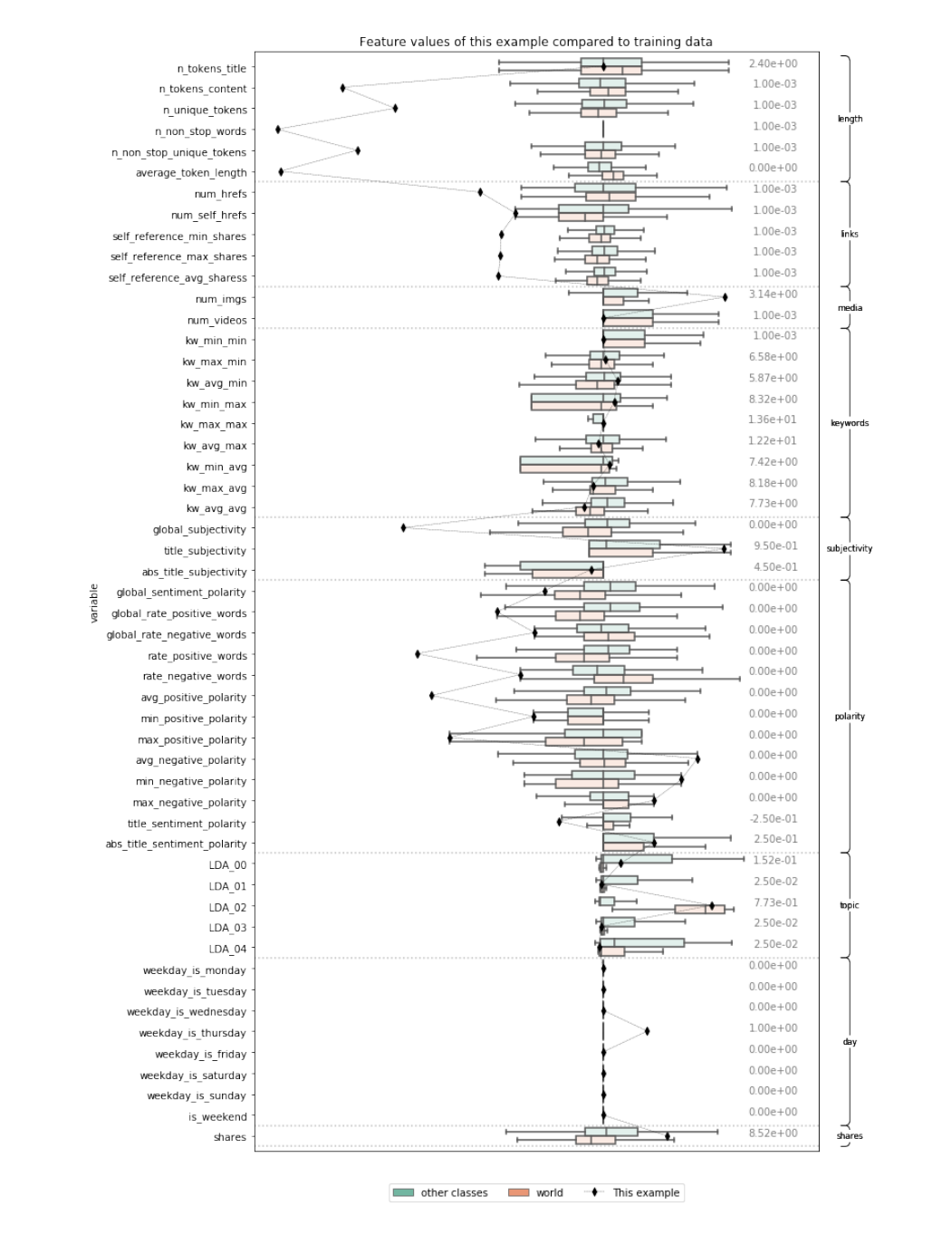}%
    \includegraphics[width=0.475\linewidth, clip, trim={1.8cm 2.5cm 2cm 2cm}, page=9]{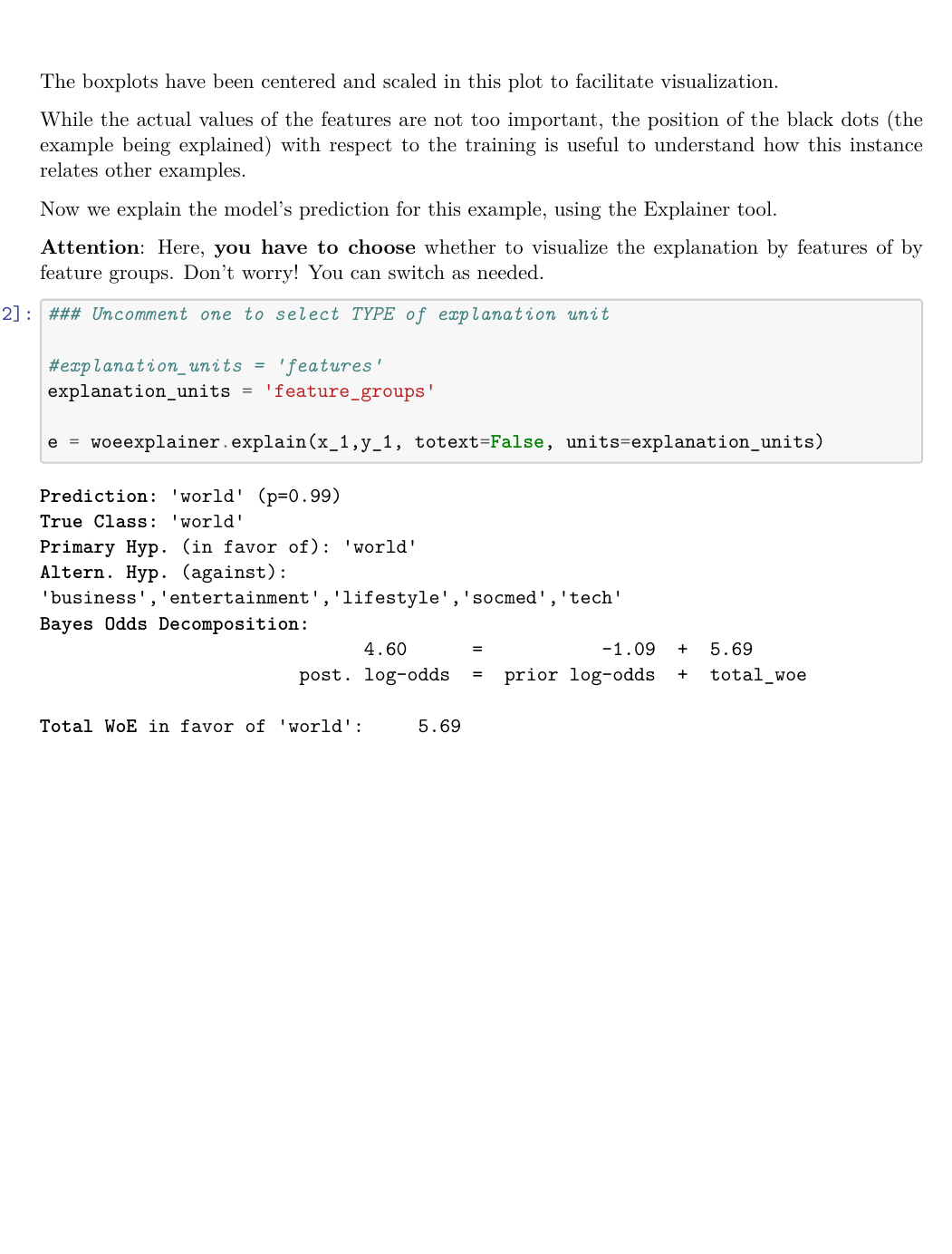}
    \includegraphics[width=0.475\linewidth, clip, trim={1.8cm 3cm 2cm 2cm}, page=10]{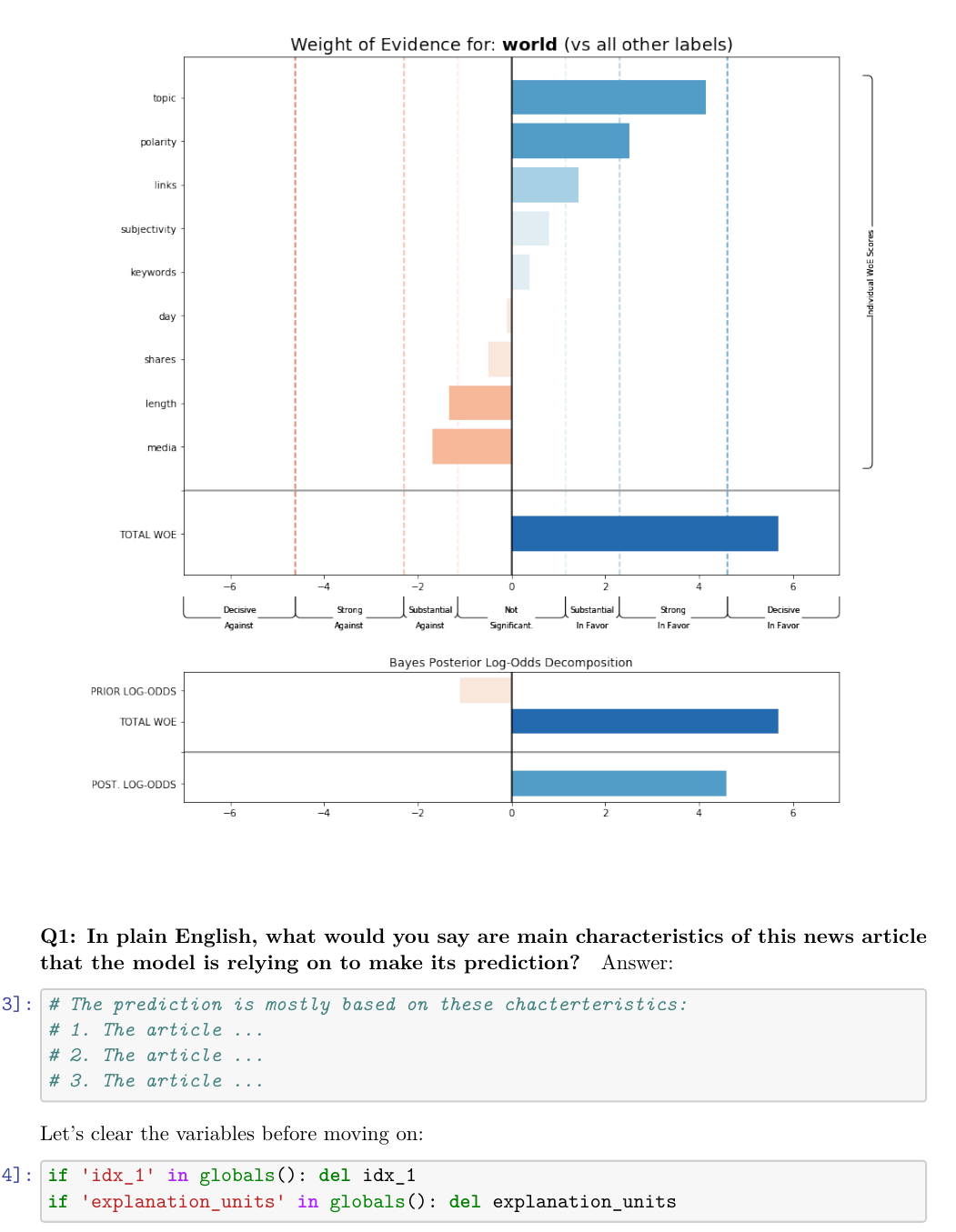}%
    \includegraphics[width=0.475\linewidth, clip, trim={1.8cm 3cm 2cm 2cm}, page=11]{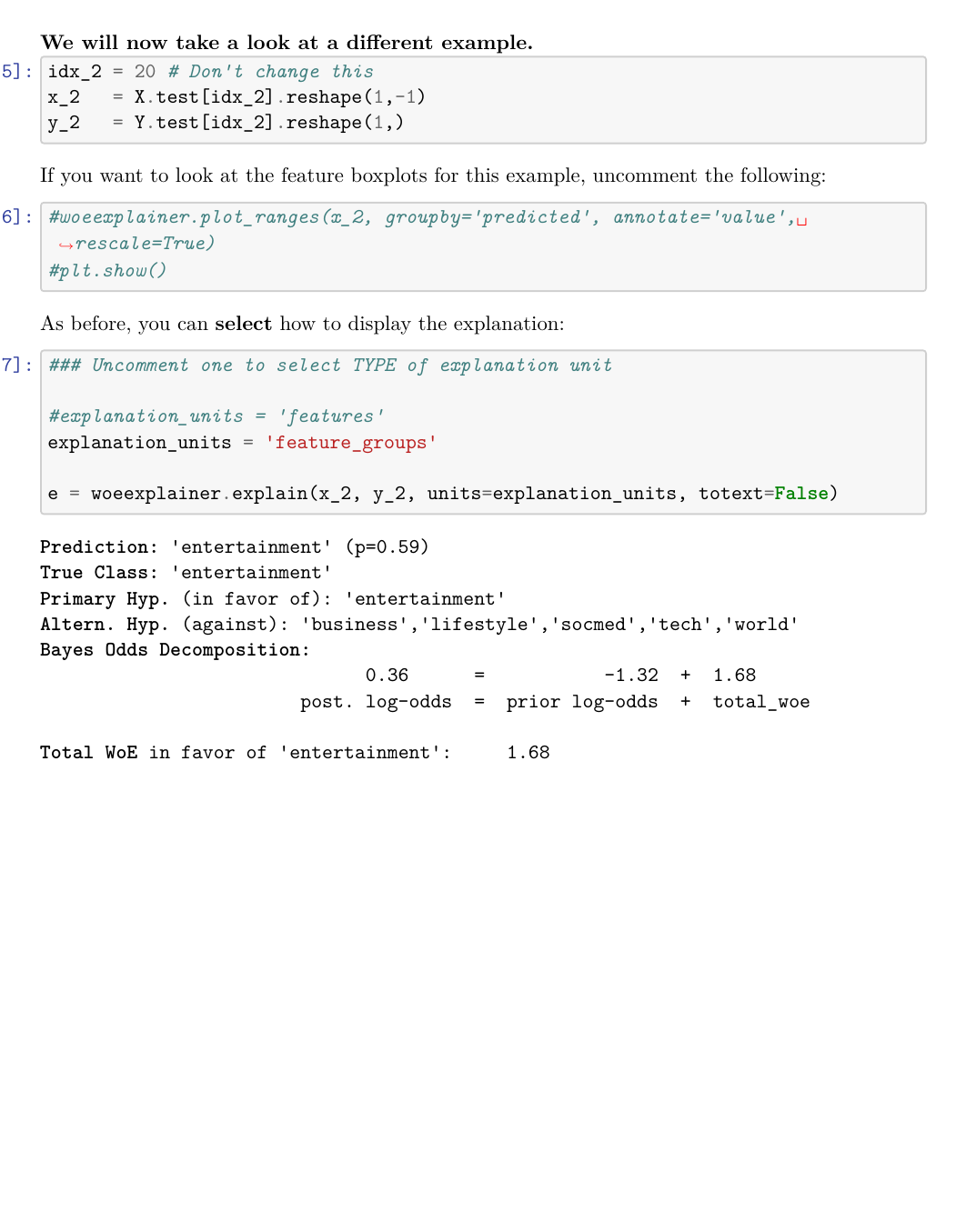}
    \caption{Jupyter notebook for the main study (continued).}
    \label{fig:study_2}
\end{figure*}

\begin{figure*}[!ht]
    \centering
    \includegraphics[width=0.475\linewidth, clip, trim={1.8cm 2.5cm 2cm 2cm}, page=12]{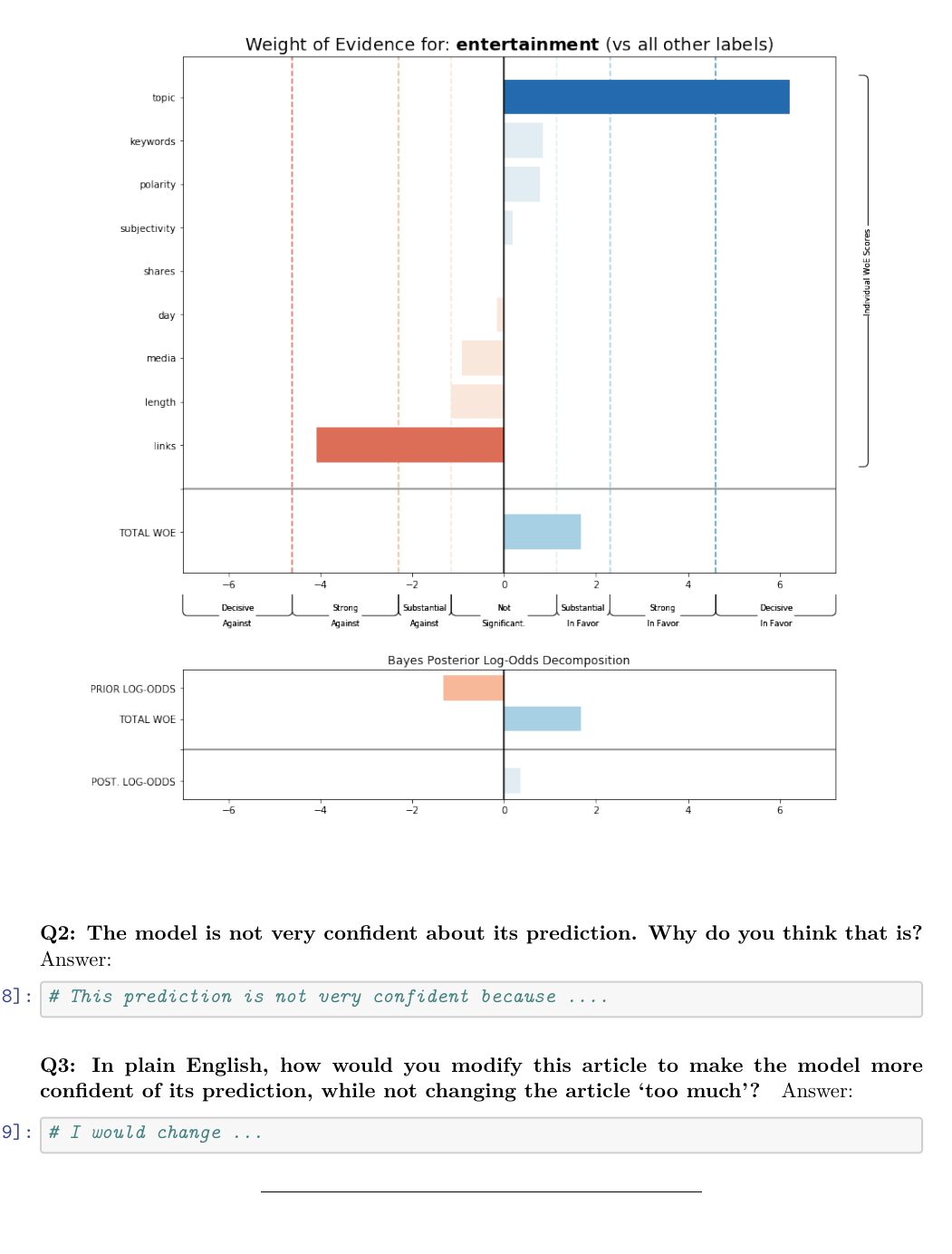}%
    \includegraphics[width=0.475\linewidth, clip, trim={1.8cm 2.5cm 2cm 2cm}, page=13]{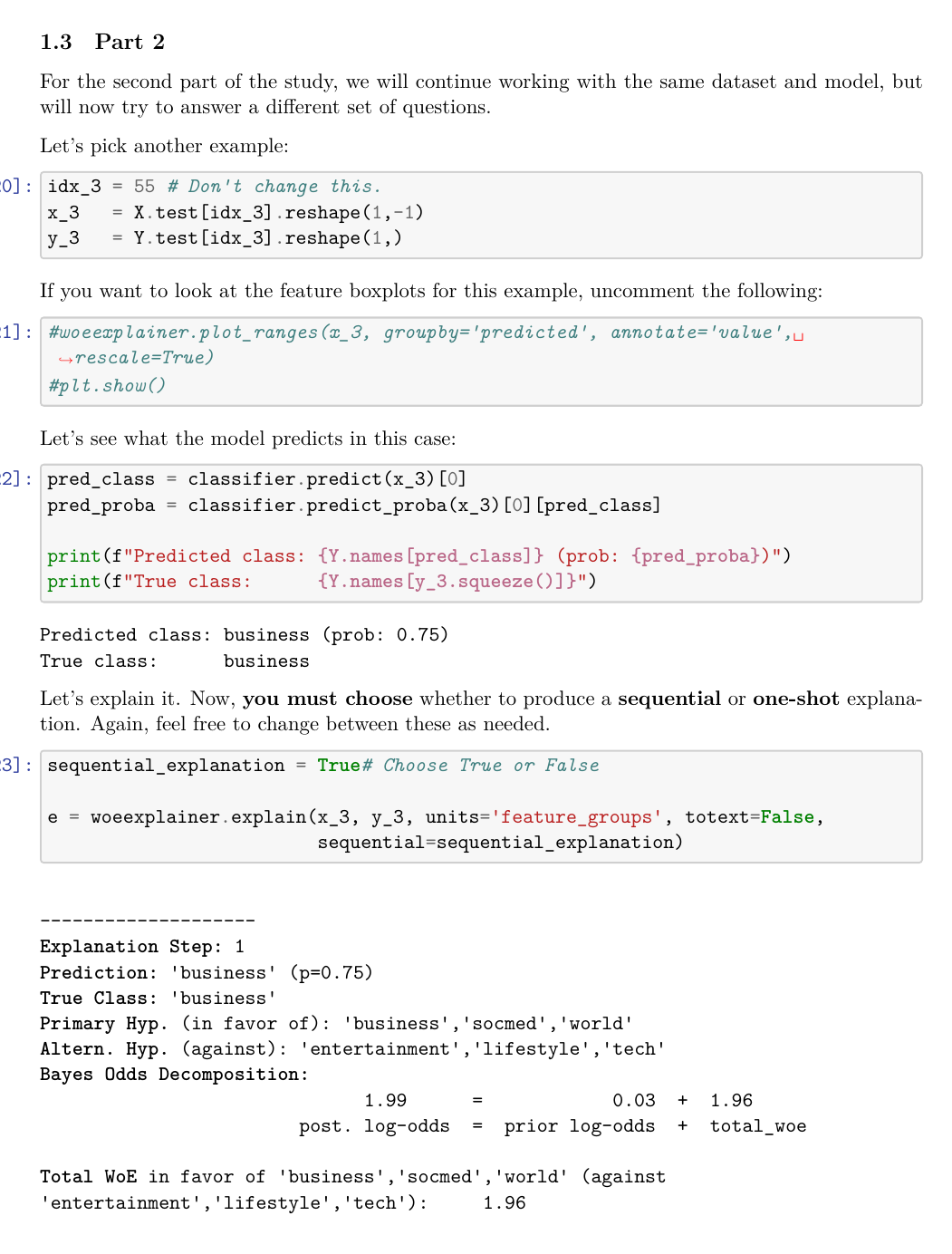}
    \includegraphics[width=0.475\linewidth, clip, trim={1.8cm 3cm 2cm 2cm}, page=14]{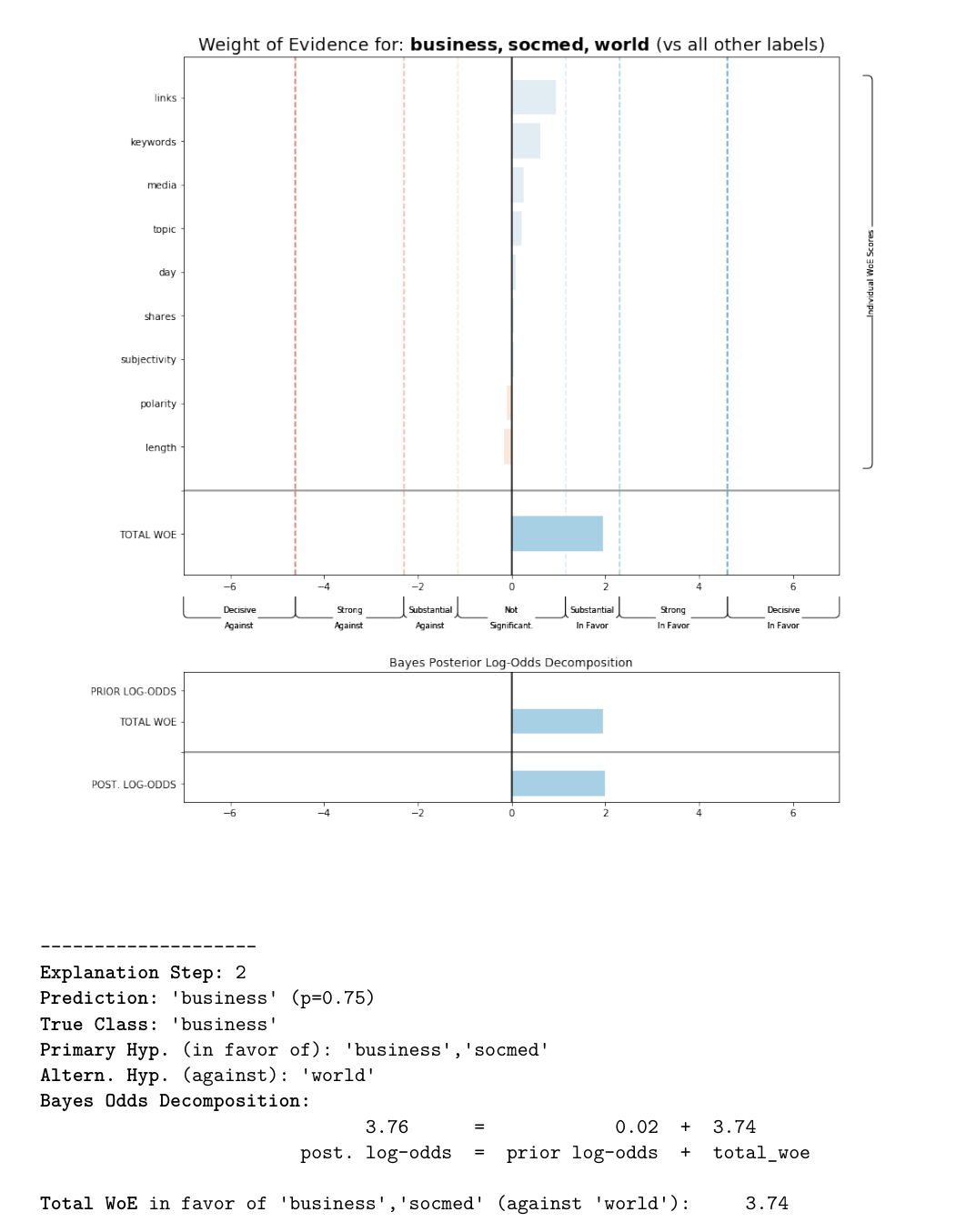}%
    \includegraphics[width=0.475\linewidth, clip, trim={1.8cm 3cm 2cm 2cm}, page=15]{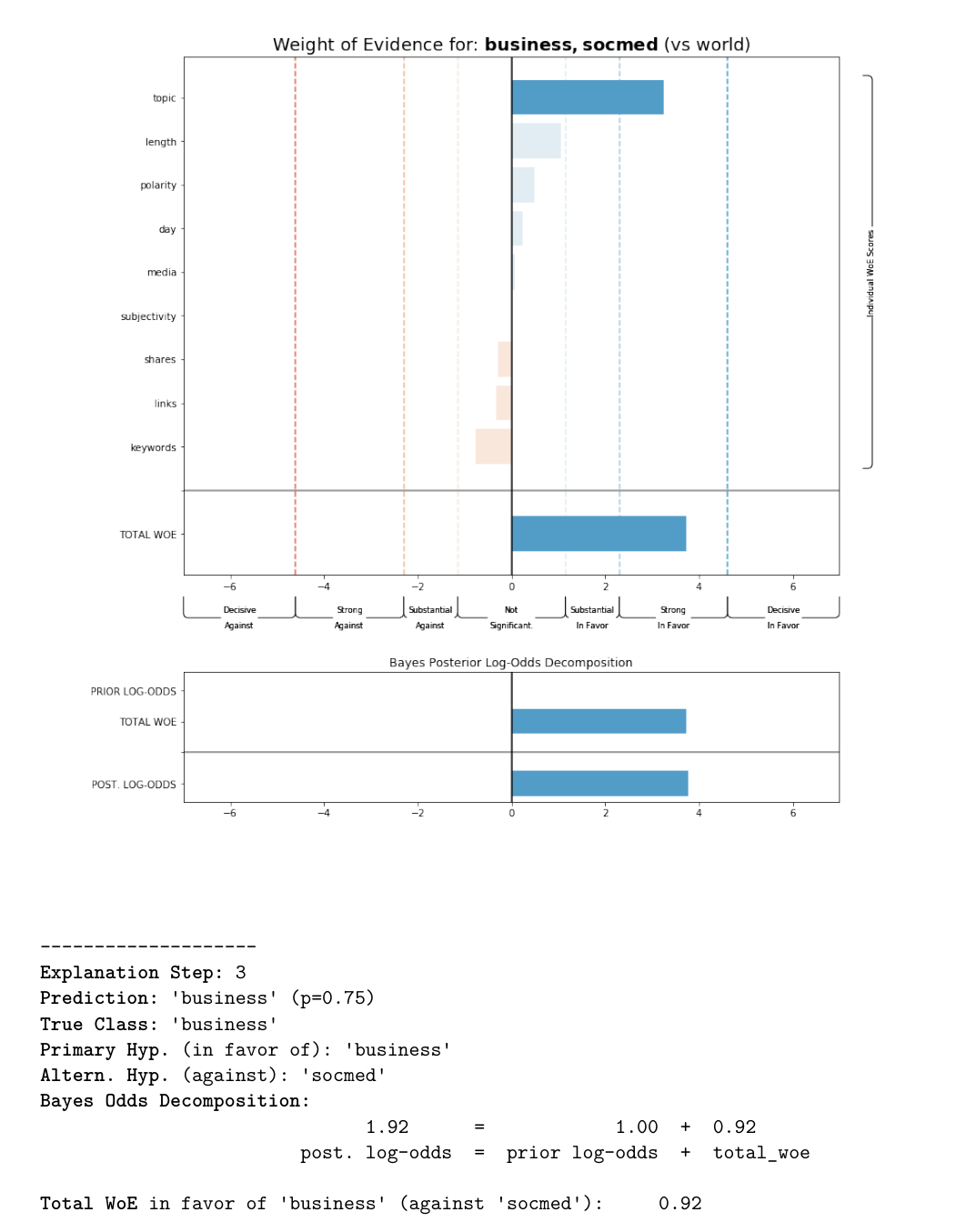}
    \caption{Jupyter notebook for the main study (continued).}
    \label{fig:study_3}
\end{figure*}

\end{document}